\documentclass[preprint,12pt,authoryear]{elsarticle}

\usepackage{amssymb}
\usepackage{amsmath}
\usepackage{framed,multirow,caption}
\usepackage{graphicx, subfigure}
\usepackage{booktabs}
\usepackage{makecell}
\usepackage{amssymb, mathrsfs}
\usepackage{threeparttable}
\usepackage{latexsym}
\usepackage{url}
\usepackage{xcolor}
\usepackage{hyperref}
\usepackage{adjustbox}
\usepackage{bbding}
\usepackage{pifont}
\usepackage[table]{xcolor}
\usepackage[ruled,noline,linesnumbered,nofillcomment]{algorithm2e}

\journal{Nuclear Physics B}

\begin{document}

\begin{frontmatter}



\title{ReSeg: Retraining-free segmentation of abdominal organs via generalizable semantic representation learning} %% Article title


\author[SJTU]{Derong Yu} %% Author name
\author[SJTU]{Guoyan Zheng}%[]

\affiliation[SJTU]{organization={Institute of Medical Robotics, School of Biomedical Engineering, Shanghai Jiao Tong University},
            addressline={No.800 Dongchuan Road}, 
            city={Shanghai},
%          citysep={}, % Uncomment if no comma needed between city and postcode
            postcode={200240}, 
            country={China}}

%% Abstract
\begin{abstract}
Abdominal organ segmentation is essential in modern medical applications including computer-aided diagnosis, therapy planning, and surgical navigation. While existing deep learning approaches demonstrate promising performance, their heavy reliance on large-scale patient image data covering multiple domains raises concerns over generalization capability and privacy risks. This motivates the development of generalizable segmentation training paradigms based on domain-randomized image synthesis, where networks are trained using synthesized images with diverse contrasts and resolutions, facilitating generalization across unseen real image domains. However, existing paradigms either suffer from limited segmentation performance when applied to abdominal organs, or require dataset-specific finetuning. To address these limitations, we propose ReSeg, a novel two-stage retraining-free framework for domain-agnostic zero-shot abdominal organ segmentation. At the first stage, our organ-agnostic synthetic data-driven pretraining approach exploits semantically and visually diverse synthetic images generated by a dedicated data engine to optimize a Generalizable Semantic Representation Network (GSRNet). The network is thereby capable of learning semantic features robust against variations in anatomical structure, imaging contrast, resolution and spatial context, which provide universal representation capabilities that serve as a foundation for segmentation learning. Subsequently, in the second stage, we continue to train an Organ-specific Encoder-decoder Segmentation Network (OESNet), whose encoder is initialized with the pretrained weights. This training process is facilitated by another dedicated data engine, which generates diversified visual images while strictly retaining the inherent spatial layout of abdominal organs. To enhance organ-relevant region perception, OESNet further incorporates a Semantics-guided Segmentation (SS) decoder, which hierarchically integrates rich semantic priors derived from the well-trained GSRNet. Comprehensive evaluation on seven abdominal organ segmentation datasets across three different modalities including CT, MRI, and PET demonstrates that ReSeg consistently outperforms existing state-of-the-art (SOTA) methods while eliminating the need for real training images. The source code will be publicly available at \url{https://github.com/Kendra9999/ReSeg}.
\end{abstract}


%% Keywords

\end{frontmatter}


\section{Introduction}

\begin{figure}[tbp]
\centering%% For centre alignment of image.
\includegraphics[width=1.0\textwidth]{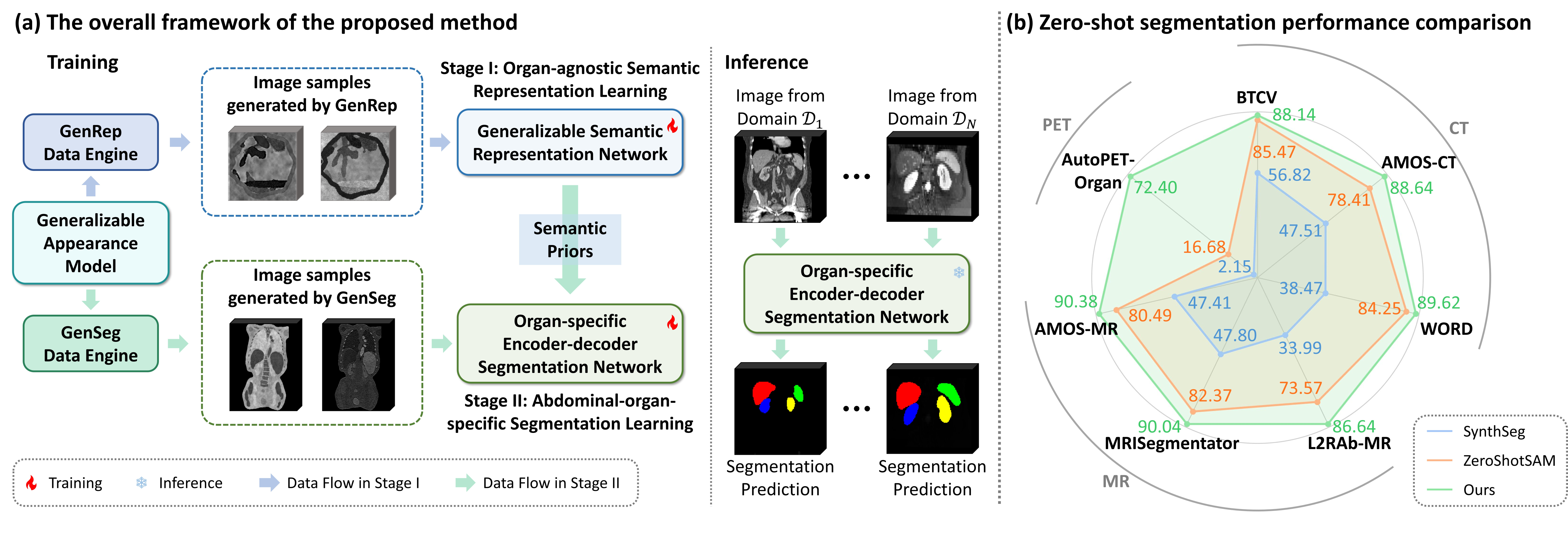}
\caption{Illustration of the proposed two-stage retraining-free framework and the zero-shot segmentation performance. (a) We utilize organ shape templates to randomly sample images from a wide range of distributions and train the model on the synthetic images through two learning stages. Stage I learns organ-agnostic, generalizable semantic features that are robust to imaging variations, while stage II leverages these semantic priors to learn abdominal-organ-specific semantic features for image segmentation. In the inference stage, our model can segment real images from multiple domains without retraining or finetuning. (b) ReSeg achieves superior performance compared with existing state-of-the-art (SOTA) methods on seven abdominal segmentation datasets across three different modalities including CT, MRI, and PET. In particular, SynthSeg \citep{billot2023synthseg} is trained only on synthetic data while ZeroShotSAM \citep{shaharabany2024zero} leverages Segment Anything Model (SAM) \citep{kirillov2023segment} to perform minimally-guided zero-shot segmentation using foreground prompt points as the input. The solid line in the radar plot depicts the average Dice Similarity Coefficient (DSC) on each dataset.}
\label{fig_motivation}
\end{figure}

Abdominal organ segmentation plays a pivotal role in medical imaging by enabling precise organ localization and volumetric analysis, serving as a foundation for computer-aided diagnosis \citep{chan2020computer}, therapy planning \citep{chen2021deep}, and surgical navigation \citep{okamoto2015clinical}. For instance, liver segmentation aids focal lesion detection and liver fibrosis staging \citep{tong2015discriminative,gao2023reliable}, and kidney segmentation from preoperative scans is a prerequisite for building 3D models that support augmented reality navigation in laparoscopic partial nephrectomy \citep{zhang2019markerless}. Manual segmentation of abdominal organs is time-consuming, labor-intensive, and error-prone, thus spurring the need for accurate automatic segmentation algorithms.

Automated segmentation of abdominal organs from volumetric medical images is a challenging task. First, soft tissues usually have low contrast characteristic in abdominal imaging, leading to weak boundaries between adjacent anatomical structures \citep{wang2019abdominal}. Second, the abdominal cavity presents a complex background with highly variable morphological structures, whose shapes and sizes exhibit significant heterogeneity among patients \citep{ma2021abdomenct}. Third, abdominal images acquired from different clinical settings often vary substantially in terms of field of view (FOV), introducing inconsistencies in the contextual information associated with organs.

The field has witnessed remarkable progress through deep learning approaches to meet the challenges mentioned above \citep{isensee2021nnu,liu2023clip,ma2024segment,zhao2025foundation}. This progress has shifted the research focus from dataset-specific models towards universal foundation models, which offer improved performance while enhancing generalization capability. Some studies leverage approaches inspired by Contrastive Language-Image Pretraining (CLIP) \citep{liu2023clip} or Segment Anything Model (SAM) \citep{du2024segvol,ma2024segment,he2025vista3d} to handle diverse segmentation tasks, while others advance representation learning methods to pretrain networks with powerful feature extraction capability. These include reconstruction-based methods like MAE \citep{he2022masked}, SimMIM \citep{xie2022simmim}, and MiM \citep{zhuang2025mim}, as well as contrastive learning approaches such as VoCo \citep{wu2024voco}, UAE \citep{bai2025uae}, and SuperCL \citep{zeng2026supercl}. Finetuning these pretrained networks on downstream segmentation datasets has yielded promising results.

Despite these advantages, most existing biomedical foundation segmentation methods face a fundamental limitation: they require large amounts of medical image datasets covering multiple domains to achieve excellent generalization performance and inherently have high reliance on real medical images for training \citep{liu2023clip,du2024segvol,ma2024segment,he2025vista3d}. Such reliance introduces not only privacy risks but also modality-specific biases, as significant domain gaps persist with large-scale data \citep{huang2025eval}, leading models to learn suboptimal feature representations. This results in limited generalization in real clinical scenarios, particularly when processing images from entirely unseen biomedical scenarios with variant imaging modalities, such as Computed Tomography (CT), Magnetic Resonance Imaging (MRI), and Positron Emission Tomography (PET). Hence, there is an urgent need for a domain-agnostic zero-shot segmentation approach, where the model can segment images from unseen domains with cross-modality discrepancies and intra-modality variations. 

Recently, there have been research efforts dedicated to developing generalizable segmentation training paradigms based on domain-randomized image synthesis strategies in brain MR image analysis \citep{billot2023synthseg,ayaz2024brain,gopinath2025recon}. These approaches train Convolutional Neural Networks (CNNs) exclusively on synthetic images sampled from a generative model conditioned on label maps, eliminating the requirement for real training images. By leveraging a domain randomization strategy, they generate pseudo training images with varying contrasts and resolutions, exposing models to extensive variability during training. This forces the models to learn domain-independent features, enabling effective generalization to real clinical scans across diverse domains during inference. Beyond brain imaging, such a paradigm has also been extended in existing literature to segmentation tasks targeting other anatomical structures, including vascular structures \citep{chollet2024label} and cellular nuclei \citep{dey2024anystar}. Building on the domain randomization concept, Anatomix \citep{dey2025learning} further introduced a general-purpose representation learning framework powered by a dedicated data engine, which synthesizes highly variable training samples from compositions of organ shape templates. This enables the pretraining of a network that serves as a strong, dataset-agnostic initialization for finetuning on new segmentation datasets.

Despite significant progress achieved so far, to achieve generalizable segmentation of abdominal organs without access to real training images, several challenges persist. First, most existing approaches are designed for structures with relatively fixed anatomical topology and stable tissue intensity distribution such as the brain \citep{billot2023synthseg}, the vascular structures \citep{chollet2024label}, and the cellular nuclei \citep{dey2024anystar}, and thus fall short in segmentation performance when applied to abdominal organs (as demonstrated by the experimental results presented in Fig. \ref{fig_motivation}), which exhibit large morphological variations among patients. Second, it is challenging to generate realistic synthetic abdominal data due to the lack of standardized anatomical atlases and refined annotated templates, resulting in severe domain gaps between synthetic and real clinical scans. Third, the existing domain-randomization-based representation learning method Anatomix \citep{dey2025learning} focuses on acquiring appearance-agnostic representations, yet overlooks the contextual consistency in the learned features. This may lead to diminished performance when segmenting images across varying FOVs where contextual information changes. Finally, although the existing representation learning method Anatomix \citep{dey2025learning} enhances domain-agnostic features, additional finetuning on downstream datasets is required for adapting the pretrained network to task-specific segmentation, limiting their applicability in retraining-free settings. 

To address these limitations, we first propose a Generalizable Appearance (GA) model to generate diversified pseudo-images based on organ shape templates, thereby eliminating the reliance on real annotated images for network training. This model realizes full randomization of image contrast and resolution via domain randomization. On this basis, we further develop two dedicated data engines, namely GenRep and GenSeg, which are respectively devised for organ-agnostic semantic representation learning and abdominal-organ-specific segmentation learning. Specifically, GenRep generates image samples with rich semantic layouts and visual variations to capture organ-agnostic, generalizable semantic representations. In contrast, GenSeg yields appearance-variant pseudo-images while retaining inherent anatomical spatial configurations. Combined with background simulation empowered by a newly designed Label Processing (LP) model, it enables optimization for abdominal organ segmentation without any real training images.

Built upon these two data engines, we propose a two-stage retraining-free framework, referred to as ReSeg, for real-training-image-free abdominal organ segmentation, as visualized in Fig. \ref{fig_motivation}-(a). Specifically, at the first stage, our organ-agnostic synthetic data-driven pretraining approach exploits semantically and visually diverse synthetic images generated by the GenRep data engine to optimize a Generalizable Semantic Representation Network (GSRNet). The network is thereby capable of learning semantic features robust against variations in anatomical structure, imaging contrast, resolution and spatial context, which provide universal representation capabilities that serve as a foundation for segmentation learning. This is done by designing a novel Multi-patch Context-aware Contrastive (MCC) loss that encourages domain-invariant feature alignment amidst contextual variations. Subsequently, in the second stage, we continue to train an Organ-specific Encoder-decoder Segmentation Network (OESNet), whose encoder is initialized with the pretrained weights. This training process is facilitated by the GenSeg data engine, which generates diversified visual images while strictly retaining the inherent spatial layout of abdominal organs. To enhance organ-relevant region perception, OESNet further incorporates a Semantics-guided Segmentation (SS) decoder, which hierarchically integrates rich semantic priors derived from the well-trained GSRNet. During inference, ReSeg can segment real clinical images from multiple domains without retraining or finetuning. ReSeg achieves superior performance compared with existing state-of-the-art (SOTA) methods on seven abdominal segmentation datasets across three different modalities (see Fig. \ref{fig_motivation}-(b)). Our primary contributions are summarized as follows:

\begin{itemize}
\item We introduce a Generalizable Appearance (GA) model that can synthesize images with diverse contrasts and resolutions from given organ shape templates. Based on this model, we further introduce two dedicated data engines, referred to as GenRep and GenSeg, respectively designed to support organ-agnostic semantic representation learning and abdominal-organ-specific segmentation learning.
\item Built upon these two data engines, we propose ReSeg, a novel two-stage retraining-free framework for real-training-image-free abdominal organ segmentation.
\item In the proposed framework, the generalizable semantic representation learning at the first stage is achieved by designing a novel Multi-patch Context-aware Contrastive (MCC) loss that encourages domain-invariant feature alignment amidst contextual variations, while the abdominal-organ-specific segmentation learning at the second stage is facilitated by the pretrained weights and by incorporating semantic priors into a Semantics-guided Segmentation (SS) decoder.  
\item Comprehensive experiments on seven abdominal organ segmentation datasets across three different modalities including CT, MRI and PET demonstrate the superior performance of our proposed ReSeg over the SOTA approaches.
\end{itemize}

\section{Related work}

\subsection{Dataset-specific models for medical image segmentation}
The unprecedented developments of deep learning have revolutionized the field of medical image segmentation, driving a paradigm shift from traditional methods \citep{wolz2013automated,tong2015discriminative} toward deep learning-based approaches \citep{isensee2021nnu,ma2022fast}. The U-shaped encoder-decoder framework, exemplified by U-Net \citep{ronneberger2015u} with skip connections and multi-scale feature extraction, has become the dominant architecture. Numerous variants have been developed to further improve segmentation accuracy for various anatomical structures, like abdominal organs \citep{ma2021abdomenct,ji2022amos}. One strategy integrates attention mechanisms to recalibrate feature responses and emphasize target regions, thereby mitigating interference from complex abdominal background and enhancing boundary delineation of ambiguous organs \citep{wang2019abdominal,zheng2025weak}. Transformer-based architectures have also been adopted to capture long-range spatial dependencies across volumetric abdominal images, complementing local CNN features with a global perspective \citep{cao2022swin,ding2026ctcsanet}. More recently, Mamba-based models have emerged as promising alternatives, offering long-range modeling with linear complexity and thus balancing performance and computational efficiency for volumetric abdominal image segmentation \citep{ruan2024vm,li2025diversity,tan2026large}. Despite these advances, these methods are typically trained on a single dataset and thus remain dataset-specific, lacking the ability to generalize across heterogeneous imaging domains.

\subsection{Self-supervised learning for medical image segmentation}
To improve model generalization capability, remarkable progress has been witnessed in developing foundation models pretrained on large-scale medical image datasets  \citep{butoi2023universeg,liu2023clip,du2024segvol,wu2024voco,ma2024segment,he2025vista3d}. A key direction within this effort involves designing self-supervised proxy tasks to enhance representation learning. These approaches typically fall into two categories. The first employs reconstructive losses to model anatomical structures and their spatial relationships, with representative works including MAE \citep{he2022masked}, SimMIM \citep{xie2022simmim}, and MiM \citep{zhuang2025mim}. The second category, by contrast, utilizes contrastive losses to learn consistent semantic representations. This is achieved by enforcing feature similarity within a predefined semantic space, as exemplified by VoCo \citep{wu2024voco}, UAE \citep{bai2025uae}, and SuperCL \citep{zeng2026supercl}. Despite significant progress, existing representation learning methods remain limited by their high reliance on heuristics tailored to their target datasets and typically learn feature similarity within a single imaging modality. This leads to compromised generalization when applied to new biomedical domains. Although UAE \citep{bai2025uae} extends feature learning to cross-modality CT-MRI scenarios, its iterative training process is computationally intensive, and the learned features remain insufficient under significant domain shifts (e.g. CT or MRI to PET). To gain domain-agnostic performance, our work introduces domain-randomization-based data engines that synthesize images with diverse resolutions and contrasts, thereby facilitating generalizable semantic representation learning.

\subsection{Domain randomization for medical image segmentation}
Domain-randomization generative models seeks to bridge all domain gaps through extreme augmentation beyond realism without acquiring real samples \citep{tobin2017domain,tremblay2018training}. These methods fully randomize the generation parameters to produce highly unrealistic samples, in which case real images are encompassed within the synthetic data spectrum \citep{tremblay2018training}. SynthSeg \citep{billot2023synthseg} pioneered this paradigm for brain segmentation by training a 3D UNet \citep{ronneberger2015u} exclusively on vastly different synthetic images sampled from a generative model conditioned on brain label maps. This facilitates the network to learn modality-agnostic features, enabling segmentation of real brain MR scans from a wide range of target domains without requiring retraining. Subsequent researches have adopted this domain-randomization-based training framework for brain scan segmentation \citep{ayaz2024brain,gopinath2025recon}, while other studies have extended its application to vascular structures \citep{chollet2024label} and cellular nuclei \citep{dey2024anystar}. 

Although these methods enable generalizable segmentation without retraining, their direct application to abdominal organ segmentation remains limited in performance. This is primarily due to several intrinsic challenges posed by abdominal anatomy. Unlike the structures targeted by existing approaches such as the brain \citep{billot2023synthseg}, the vascular structures \citep{chollet2024label}, and the cellular nuclei \citep{dey2024anystar}, which exhibit relatively fixed anatomical topology and stable tissue intensity distribution, abdominal organs reside in a complex background and show large morphological variations among patients. This problem is further compounded by the lack of standardized anatomical atlases and refined annotated templates for the abdomen, making it difficult to generate realistic synthetic abdominal data. In contrast, our method is specifically designed for real-training-image-free abdominal organ segmentation by incorporating a generalizable appearance model and a label processing model into the pseudo-image synthesis pipeline, thereby enabling the network to effectively learn abdominal-organ-specific semantic features.

To learn general-purpose features, Anatomix \citep{dey2025learning} proposed a representation learning framework driven by a dedicated data engine that randomizes spatial configurations of organ shape templates to synthesize paired image samples with distinct appearances. During pretraining, a contrastive learning strategy is employed to encourage a network to be stable against nuisance imaging variations that do not change image semantics. The pretrained network can then be used to extract robust, domain-agnostic representations for downstream tasks. Nevertheless, this method primarily focuses on appearance invariance while neglecting contextual consistency, which is crucial for segmenting images with contextual changes across varying FOVs. Additionally, finetuning on downstream datasets is required for image segmentation, which may introduce complexity for clinical users. To bridge these gaps, we propose a two-stage retraining-free framework that first learns organ-agnostic semantic features robust to appearance and contextual changes, and then transitions to learning organ-specific semantic features with the guidance of these semantic priors.

\begin{figure}[tbp]
\centering%% For centre alignment of image.
\includegraphics[width=1.0\textwidth]{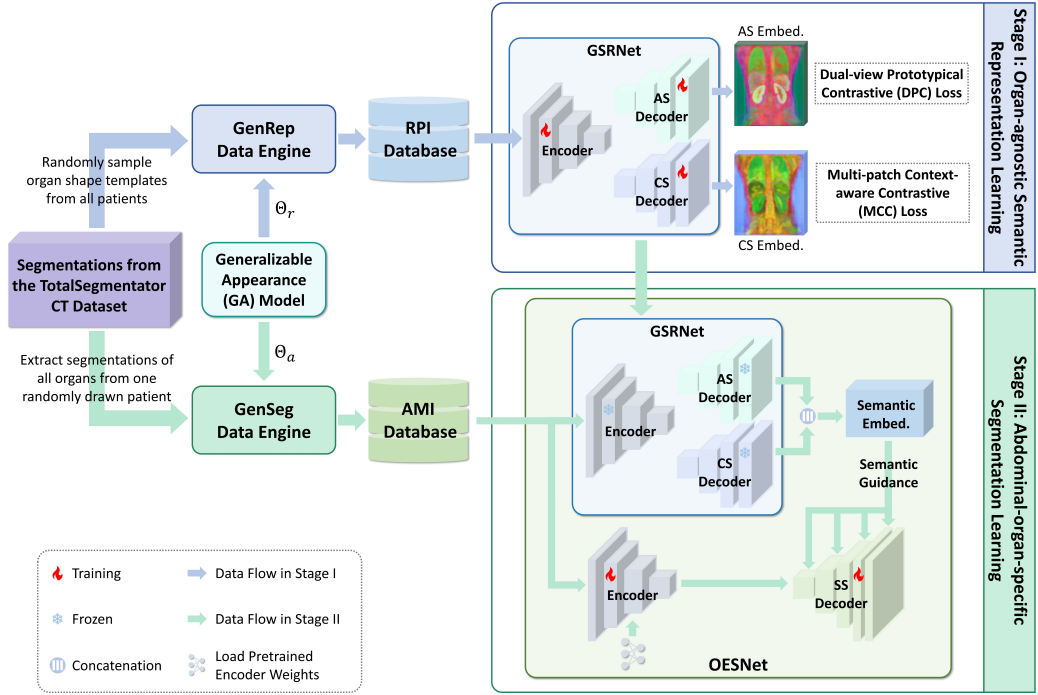}
\caption{A schematic overview of ReSeg. We leverage segmentation masks from the TotalSegmentator CT dataset~\citep{wasserthal2023totalsegmentator}, excluding real images entirely. Two data engines, referred to as GenRep and GenSeg, are introduced to construct a Randomized Paired Image (RPI) database and an Anatomical Multi-view Image (AMI) database, respectively. ReSeg is a two-stage learning framework that is trained on these two databases. In particular, in stage I, the RPI database is used to train a Generalizable Semantic Representation Network (GSRNet), which learns organ-agnostic, generalizable semantic features with an Appearance-aware Semantic (AS) decoder and a Context-aware Semantic (CS) decoder. In stage II, an Organ-specific Encoder-decoder Segmentation Network (OESNet) is initialized with the pretrained encoder weights and then trained on the AMI database. OESNet is guided by semantic priors calculated by the well-trained GSRNet to enhance domain-invariant anatomical feature extraction within a Semantics-guided Segmentation (SS) decoder. We reduce the dimension of the embeddings to 3 using PCA and visualize them as RGB images. Embed.: embedding.}
\label{fig_method_overview}
\end{figure}

\section{Method}

\subsection{Overview}
This work aims to enable model generalization across diverse domains without acquiring any real training images. For training, we access a publicly available database of segmentation masks from the TotalSegmentator CT dataset \citep{wasserthal2023totalsegmentator}, which encompasses 1,204 patients with annotations for 104 anatomical structures. Notably, real images are entirely excluded, and only organ shape templates derived from the segmentation masks are adopted. Based on these organ shape templates, ReSeg employs a two-stage pipeline:

\textbf{1) Pretraining stage:} A Generalizable Semantic Representation Network (GSRNet) is pretrained to learn organ-agnostic, generalizable semantic features. As shown in Fig. \ref{fig_method_overview}, the training process relies on a Randomized Paired Image (RPI) database, which is constructed by a delicately designed GenRep data engine and provides paired pseudo-images with rich semantic layouts and diverse appearances. To capture semantic features robust against appearance and contextual variations, two specialized decoders are employed: an Appearance-aware Semantic (AS) decoder optimized via a Dual-view Prototypical Contrastive (DPC) loss that promotes domain-invariant feature learning, and a Context-aware Semantic (CS) decoder trained with a Multi-patch Context-aware Contrastive (MCC) loss that encourages feature alignment amidst contextual variations. The learned features offer universal representation capabilities, serving as a foundation for subsequent segmentation learning.

\textbf{2) Segmentation learning stage:} We proceed to train an Organ-specific Encoder-decoder Segmentation Network (OESNet), which leverages the semantic priors acquired in the pretraining stage to learn abdominal-organ-specific semantic features. As depicted in Fig. \ref{fig_method_overview}, OESNet is initialized with the pretrained encoder weights and then fine-trained on an Anatomical Multi-view Image (AMI) database. This database is generated by another delicately designed GenSeg data engine that synthesizes appearance-variant image samples while preserving inherent anatomical spatial configurations. To enhance organ-relevant region perception, OESNet further incorporates a Semantics-guided Segmentation (SS) decoder. This decoder enhances segmentation feature extraction by hierarchically integrating rich semantic priors derived from the well-trained GSRNet.

During inference, OESNet achieves broad generalization across multiple image domains, effectively handling diverse modalities with varying acquisition conditions including vendors, scanning protocols, imaging sequences, and patient populations. Organ segmentations for real images are achieved without requiring domain-specific adaptation (e.g., retraining or finetuning). 

Below we will present the details of the two data engines and the two-stage learning framework.

\begin{figure}[tbp]
\centering%% For centre alignment of image.
\includegraphics[width=1.0\textwidth]{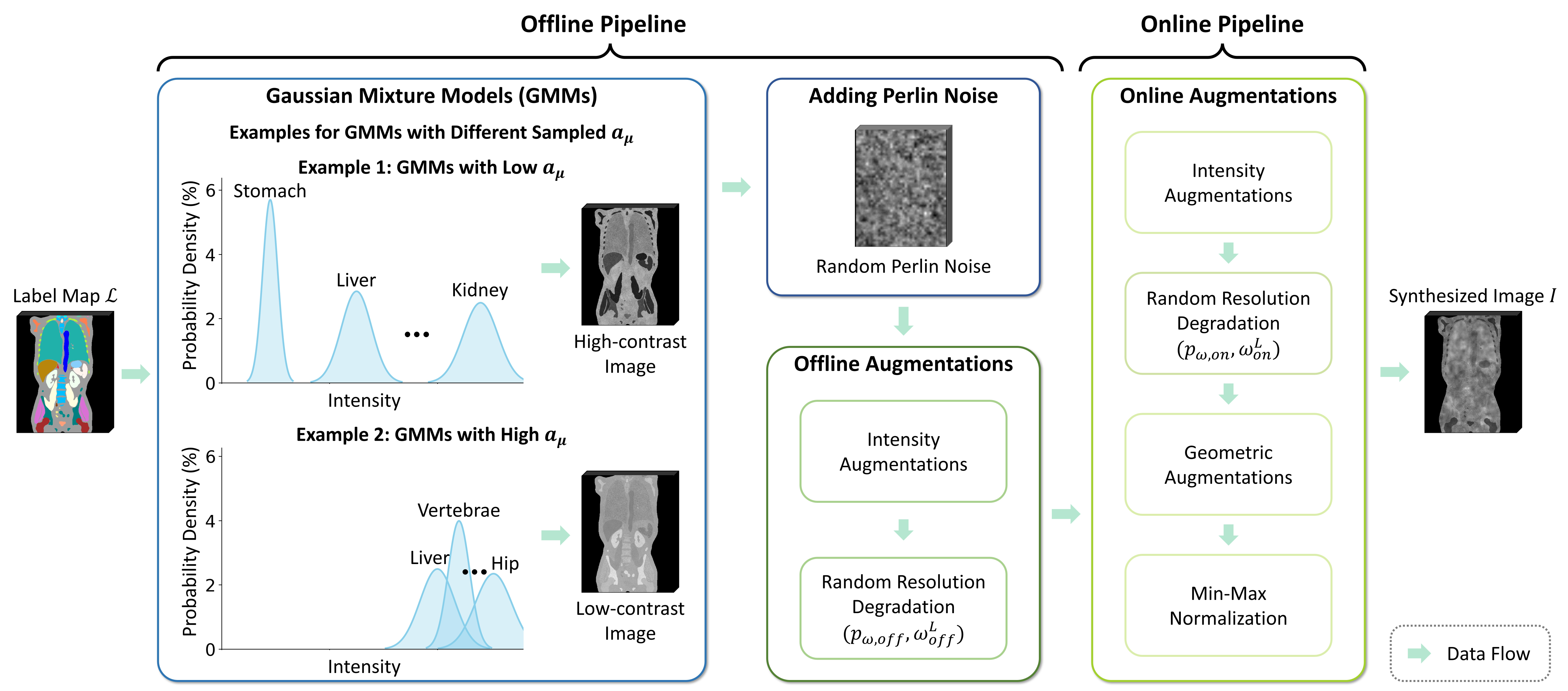}
\caption{Illustration of the Generalizable Appearance (GA) model. Given a label map $\mathcal{L}$ with $L$ labels, GMMs are used to randomly sample intensities. For all label $l\in\{1,2,\cdots,L\}$, a lower bound $a_{\mu}$ for the mean parameters are first drawn from a uniform distribution $U(a_{\mu}^L,a_{\mu}^H)$, where $a_{\mu}^L$ and $a_{\mu}^H$ are two fixed hyperparameters controlling respectively the lowest and the highest value for this dynamic lower bound. Then, for each label $l$ in $\mathcal{L}$, the intensity $I_l$ is sampled from a Gaussian distribution $\mathcal{N}(\mu_l, \sigma_l^2)$, where $\mu_l\sim U(a_{\mu},b_{\mu})$, $\sigma_l\sim U(a_{\sigma},b_{\sigma})$, and $b_{\mu}$, $a_{\sigma}$, and $b_{\sigma}$ are fixed hyperparameters. After GMMs, Perlin noise \citep{perlin1985image} is applied, followed by random offline and online augmentations.}
\label{fig_intensity}
\end{figure}

\subsection{Data engines}
Our pipeline only leverages organ shape templates to synthesize diverse pseudo training images. To this end, we propose a Generalizable Appearance (GA) model that generates pseudo-images conditioned on label maps derived from the available templates. This model employs domain randomization to fully randomize image contrasts and resolutions. Different from existing methods \citep{billot2023synthseg,dey2025learning}, we introduce a dynamic lower bound strategy in the Gaussian Mixture Models (GMMs) for intensity sampling. This strategy substantially enhances variation in image contrast (see Fig. \ref{fig_intensity}) and boosts the asymptotic scaling of contrast variance (see \textit{Supplement A} for detailed mathematical proof), thereby increasing the diversity of synthetic images to improve model generalization capability. 

Built upon the GA model, we designed two dedicated data engines based on different parameter sets and distinct label maps to respectively construct two synthetic databases, as illustrated in Fig. \ref{fig_method_overview}. The first engine, referred to as GenRep, generates image samples with diverse semantic layouts and appearances for organ-agnostic semantic representation learning. It first creates randomized label maps from randomly sampled shape templates across all patients. This randomization strategy increases semantic layout diversity, thereby enabling effective representation learning. Each randomized label map is then used to synthesize paired images with distinct appearances using the GA model, simulating cross-domain data. These randomized label maps with their paired synthetic images form the RPI database.

The second engine, namely GenSeg, is designed for abdominal-organ-specific segmentation learning. It produces image samples with varying visual appearances while preserving the inherent spatial layout of abdominal organs. Anatomical label maps are derived from the segmentation masks of all organs from the same patients, while maintaining the spatial arrangements of anatomical structures. Additionally, background simulation is integrated through a Label Processing (LP) model to support effective feature learning for abdominal organ segmentation. We then sample intensities based on these anatomical label maps using the GA model, resulting in the AMI database.

\subsubsection{Generalizable Appearance (GA) model}

\begin{algorithm}[t]
    \caption{Generalizable Appearance (GA) model}
    \label{alg_intensity}
    \SetKwProg{Offl}{Offline pipeline:}{}{}
    \SetKwProg{Onl}{Online pipeline:}{}{}
    
    \KwData{Label maps $\{\mathcal{L}\}$}
    \KwResult{Synthesized images $\{(\{I\}|\mathcal{L})\}$}
    \Offl{}{
        \For{$\mathcal{L}~{\rm in}~\{\mathcal{L}\}$}{
            \For{$i=1~{\rm to}~N_I$}{
                $a_{\mu}\sim U(a_{\mu}^L,a_{\mu}^H)$\;
                \For{$l=1~{\rm to}~L$}{
                    $\mu_l\sim U(a_{\mu},b_{\mu})$; $\sigma_l\sim U(a_{\sigma},b_{\sigma})$\;
                    $I_l \sim \mathcal{N}(\mu_l, \sigma_l^2)$\;
                }
                Apply Perlin noise\;
                Random offline augmentations\;
            }
         } 
     }
     \Onl{}{
         \While{not complete sampling for training}{
             Sample $\mathcal{L}$ and offline $I$\;
             Random online augmentations\;
         }     
     }
\end{algorithm}

Inspired by the appearance model as introduced by \cite{dey2025learning}, the GA model processes a label map $\mathcal{L}$ with $L$ distinct labels through two stages: an offline pipeline generating $N_I$ synthetic images $\{I\}$, followed by an online pipeline to further diversify them (see Algorithm \ref{alg_intensity}).

\textbf{1) Offline pipeline:} The offline pipeline begins by using GMMs with the dynamic lower bound strategy for random intensity sampling. For each synthetic image, a dynamic lower bound $a_{\mu}$ for the mean parameters is first randomly drawn from a uniform distribution $U(a_{\mu}^L, a_{\mu}^H)$, where $a_{\mu}^L$ and $a_{\mu}^H$ are two fixed hyperparameters controlling respectively the lowest and the highest value for $a_{\mu}$. After that, each label $l\in\{1,2,\cdots,L\}$ is assigned a Gaussian model $\mathcal{N}(\mu_l, \sigma_l^2)$ with the mean parameter $\mu_l\sim U(a_{\mu}, b_{\mu})$ and the standard deviation parameter $\sigma_l\sim U(a_{\sigma}, b_{\sigma})$ to sample intensity $I_l$. Here, $a_{\mu}$ is the dynamic lower bound for $\mu_l$ as described before, $a_{\sigma}$ serves as the lower bound for $\sigma_l$, and $b_{\mu}$ and $b_{\sigma}$ denote the upper bounds for $\mu_l$ and $\sigma_l$, respectively. Throughout our experiments, $a_{\sigma}$, $b_{\mu}$, and $b_{\sigma}$ are fixed hyperparameters.

The offline pipeline then applies Perlin noise \citep{perlin1985image} to emulate spatial texture and uses randomly sampled offline augmentations relevant to medical images, including bias field, Fourier-spike noise, Gamma shifts, Gaussian blurring, Gibbs ringing noise, and Gaussian sharpening. To enable networks learn features robust against resolution variations, low-resolution conditions are simulated by random resolution degradations with probability $p_{\omega,off}$. Specifically, for a pseudo-image of size $H\times W\times D$ (height, width, depth), it is interpolated to size $\lfloor\omega_H \cdot H\rfloor\times \lfloor\omega_W \cdot W\rfloor\times \lfloor\omega_D \cdot D\rfloor$ using nearest neighbor method and then trilinearly interpolated back to its original size. The factors $\omega_H$, $\omega_W$ and $\omega_D$ are individually sampled from a uniform distribution $U(\omega_{off}^L, 1)$, where $\omega_{off}^L\in (0,1)$ is the lower bound. ``$\lfloor\cdot\rfloor$'' denotes the rounding operation. This strategy generates diverse image resolutions, enhancing robustness against challenging conditions like MR scans where one axis may have extremely low resolution while others retain relatively high resolutions.

\textbf{2) Online pipeline:} To further increase diversity, the online pipeline is performed by applying randomly sampled online augmentations during training. These include Gaussian noise, Gaussian blurring, resolution degradations (with distinct probability parameter $p_{\omega,on}$ and lower bound parameter $\omega_{on}^L$), elastic deformations, affine transformations, and axis flipping. Finally, before being fed into the networks, the synthetic images undergo Min-Max normalization to standardize intensities within the $[0,1]$ interval. 

\textbf{3) Parameterization:} The GA model is defined by its parameter set $\Theta=\{N_I,a_{\mu}^L,a_{\mu}^H,a_{\sigma},b_{\mu},b_{\sigma},p_{\omega,off},\omega_{off}^L,p_{\omega,on},\omega_{on}^L\}$. By changing the parameter sets and the label maps, we can design different data engines.

\begin{figure}[tbp]
\centering%% For centre alignment of image.
\includegraphics[width=1.0\textwidth]{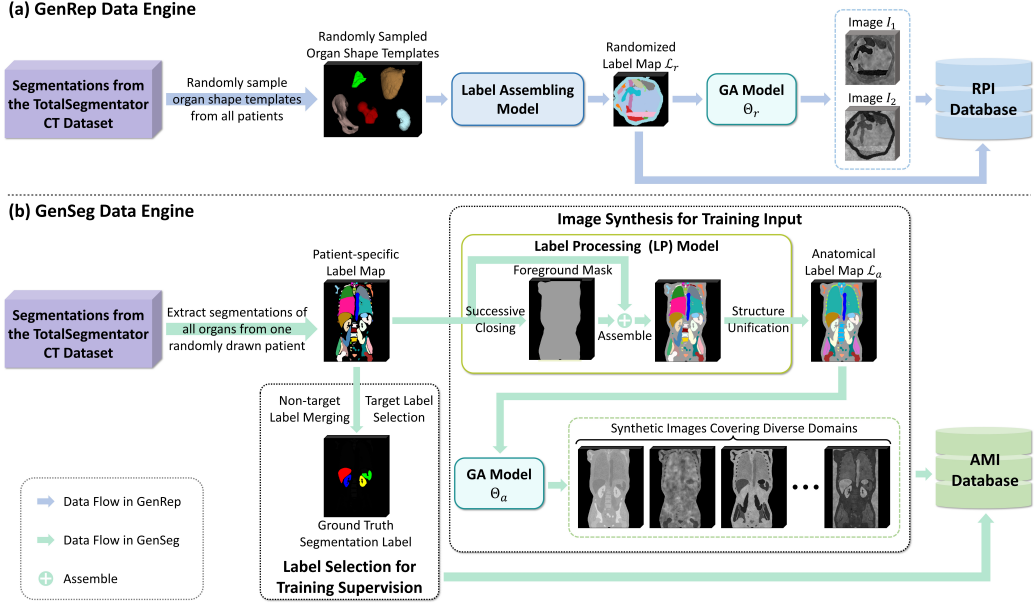}
\caption{Illustration of the two data engines for ReSeg. Both data engines use segmentation masks sampled from the TotalSegmentator CT dataset~\citep{wasserthal2023totalsegmentator}. The GenRep data engine randomly samples organ shape templates from all patients to create randomized label maps with paired synthetic images, forming a Randomized Paired Image (RPI) database. In contrast, the GenSeg data engine extracts segmentations of all organs from one randomly drawn patient to synthesize pseudo-images while preserving the inherent spatial layout of abdominal organs, constituting an Anatomical Multi-view Image (AMI) database.}
\label{fig_data_engine}
\end{figure}

\subsubsection{Data engine for representation learning}
GenRep is designed for organ-agnostic semantic representation learning. This data engine synthesizes randomized paired images with a wide range of semantic layouts and appearances, as illustrated in Fig. \ref{fig_data_engine}-(a). Following the pipeline introduced by \cite{dey2025learning}, randomized label maps are constructed by randomly sampling a random number of organ shape templates from the segmentation masks across all patients. Subsequently, a label assembling model applies random spatial deformations to the shape templates and then composite them together. The assembled label map is cropped or padded to a fixed size of $128\times128\times128$ voxels. 
 
For each randomized label map $\mathcal{L}_r$, the intensities of two paired images $I_1$ and $I_2$ are sampled from a GA model parameterized by $\Theta_{r}$. This configuration employs parameter values shown in \textit{Supplement B.1}, following established practices as introduced in \cite{dey2025learning} to facilitate generalizable representation learning. Consequently, $I_1$ and $I_2$ exhibit distinct appearances while sharing an identical semantic layout, simulating aligned images from two different domains, as shown in Fig. \ref{fig_data_engine}-(a). 

All the assembled randomized label maps and their paired images form the RPI database (refer to \textit{Supplement B.2} for data scale).

\subsubsection{Data engine for segmentation learning}

GenSeg is designed to generate diverse pseudo-images that retain inherent anatomical spatial configurations for abdominal-organ-specific segmentation learning. For a randomly drawn patient, segmentations of all organs $\{\mathcal{S}_{organ}\}$ are assembled to get a patient-specific label map while preserving the inherent spatial layout of abdominal organs (see Fig. \ref{fig_data_engine}-(b) for an example). Then, we generate pseudo-data through two steps: 1) image synthesis for training input and 2) label selection for training supervision.

\textbf{1) Image synthesis:} This step generates inputs for segmentation training. A Label Processing (LP) model is delicately designed to emulate the abdominal cavity and to create a label map for domain-randomization-based intensity sampling. Concretely, a coherent foreground mask encompassing all organs is constructed by a successive morphological closing procedure (see Fig. \ref{fig_data_engine}-(b) for an illustration). This process involves $(B+1)$ iterative closing operations defined through the recursive formulation:
\begin{equation}
\phi_b (\mathcal{S}) = 
\begin{cases}
\mathcal{S} \bullet (r_0 \mathcal{K}), & b=0 \\
\phi_{b-1} (\mathcal{S}) \bullet (r_b \mathcal{K}), & b\geq1
\end{cases}
\label{eq_closing}
\end{equation}
where $\mathcal{S}=\cup\{\mathcal{S}_{organ}\}$ represents the unified binary mask obtained by aggregating all individual organ segmentation masks. The operation ``$\bullet$'' denotes the morphological closing \citep{haralick1987image}, using a 3D spherical structuring element $\mathcal{K}$ with dynamically adjusted radius $r_b$ (in voxels). Here $b$ indexes the closing operation. The radius progression scheme for successive operations is designed as:
\begin{equation}
r_b = 2B - 2b + 1,  0 \leq b \leq B
\label{eq_radius}
\end{equation}
This parametric configuration ensures progressive refinement of mask connectivity, where the initial operation ($b=0$) employs the largest radius to establish global connectivity, followed by linearly decreased radii to preserve fine anatomical details. Through empirical validation, we fix the number of iterations at $B=6$.

The pseudo-foreground mask and the patient-specific label map are then assembled, as depicted in Fig. \ref{fig_data_engine}-(b). To reflect clinical imaging characteristics, we unify similar anatomical structures: 1) Bilateral homologous structures receive identical labels; 2) All ribs are merged into a single osseous label; 3) Vertebral column components (vertebrae + sacrum) are merged as a spinal label. After that, we obtain the anatomical label map $\mathcal{L}_a$ with $L_a$ labels.

A GA model parameterized by $\Theta_{a}$ subsequently samples diverse pseudo-images conditioned on $\mathcal{L}_a$. To capture large variations of contrasts and resolutions in abdominal imaging, we empirically configure $\Theta_{a}$ with the values shown in \textit{Supplement B.1}. This process generates the anatomical multi-view images with diverse contrasts and resolutions, as shown in Fig. \ref{fig_data_engine}-(b). 

\textbf{2) Label selection:} Supervision signals for network training are obtained by label merging and selection. In this study, we focus on segmenting four clinically significant abdominal organs: the liver, the spleen, the right kidney, and the left kidney. Thus, to generate the Ground Truth (GT) segmentation label, we process the patient-specific label map by merging all annotations corresponding to non-target structures into the background class and assigning consecutive labels to the target organs.

The GT segmentation labels and their corresponding synthetic images from all patients are compiled to form the AMI database (refer to \textit{Supplement B.2} for data scale).

\begin{figure}[tbp]
\centering%% For centre alignment of image.
\includegraphics[width=\textwidth]{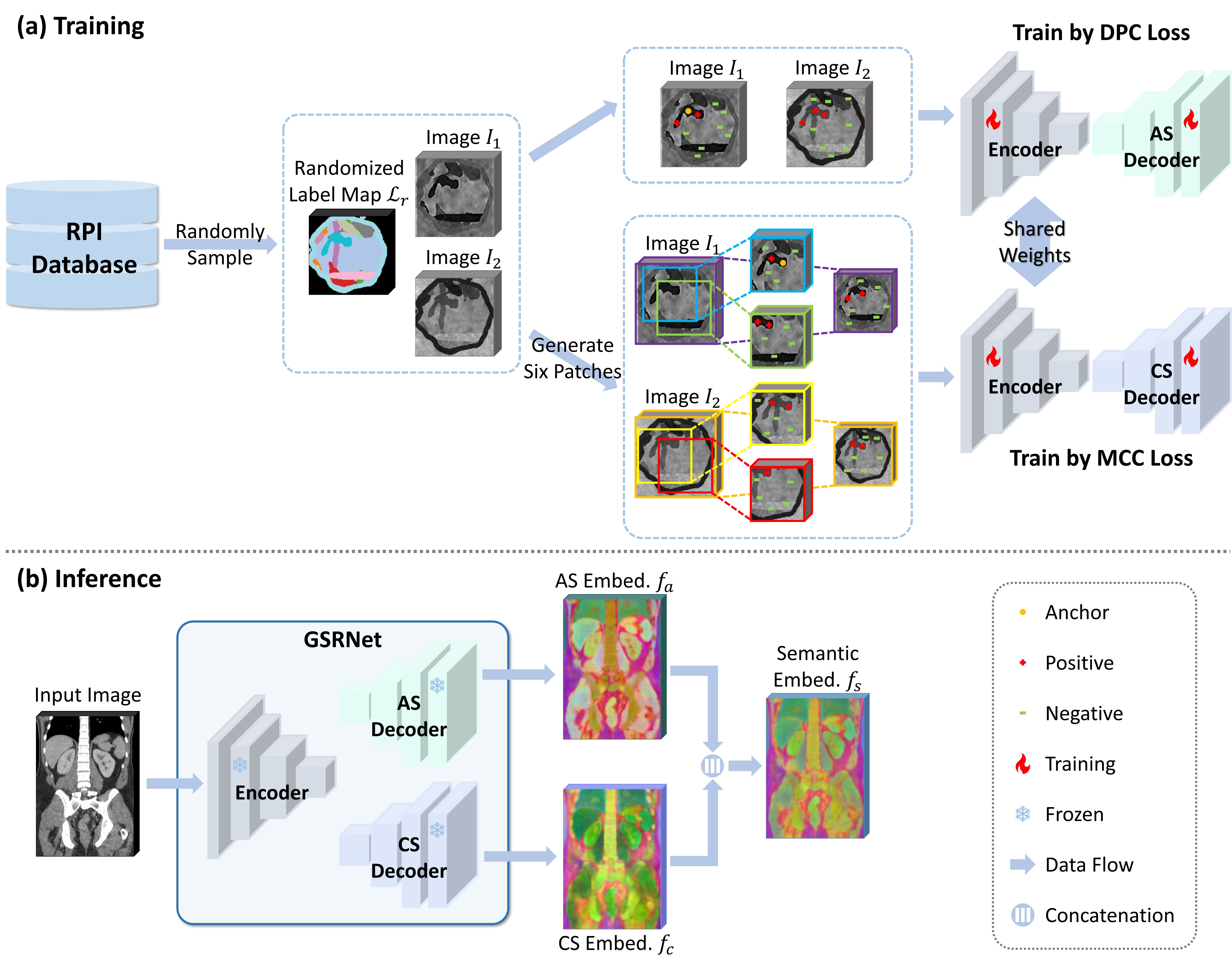}
\caption{Framework for the organ-agnostic semantic representation learning stage. During training, a synthetic image pair is randomly sampled from the RPI database to train a Generalizable Semantic Representation Network (GSRNet). This network consists of a shared CNN encoder with two distinct decoders: an Appearance-aware Semantic (AS) decoder optimized via a Dual-view Prototypical Contrastive (DPC) loss, and a Context-aware Semantic (CS) decoder trained by a Multi-patch Context-aware Contrastive (MCC) loss. During inference, an input image is fed into GSRNet to generate the AS embedding and the CS embedding from the AS decoder and the CS encoder, respectively. These embeddings are then concatenated to produce the generalizable semantic embedding. We reduce the dimension of the embeddings to 3 using PCA and visualize them as RGB images. Embed.: embedding.}
\label{fig_genrep_stage}
\end{figure}

\subsection{Organ-agnostic semantic representation learning}
This stage employs the RPI database to pretrain a Generalizable Semantic Representation Network (GSRNet). To learn semantic features robust against contextual and appearance changes, GSRNet adopts a dual-branch design, comprising a shared CNN encoder with two distinct decoders: a Context-aware Semantic (CS) decoder and an Appearance-aware Semantic (AS) decoder. Both branches produce voxel-wise semantic embeddings.

\subsubsection{Training}

\begin{algorithm}[t]
    \caption{Organ-agnostic semantic representation learning}
    \label{alg_genrep}
    \SetKwProg{CA}{CS decoder:}{}{}
    \SetKwProg{AA}{AS decoder:}{}{}
    \SetKwProg{Op}{Joint optimization:}{}{}
    
    \KwData{RPI database $\{(I_1,I_2|\mathcal{L}_r)\}$}
    \KwResult{Pretrained GSRNet (encoder $\mathcal{F}_e$, CS decoder $\mathcal{F}_c$, AS decoder $\mathcal{F}_a$)}
    \While{not complete all training iterations}{
        \CA{}{
              Sample $(I_1,I_2|\mathcal{L}_r)$\;
              Generate four patches by cropping $(I_1,I_2)$\;
              Generate two patches by resizing $(I_1,I_2)$\;
              Forward $\mathcal{F}_e$ and $\mathcal{F}_c$\;
              Randomly select $N_{anch}$ points from the overlapping region\;
              Randomly select $N_{can}$ points from each patch\;
              Calculate $\mathcal{L}_{MCC}$ by $\mathcal{L}_r$\;
        }
        \AA{}{
              Sample $(I_1,I_2|\mathcal{L}_r)$\;
              Forward $\mathcal{F}_e$ and $\mathcal{F}_a$\;
              Calculate $\mathcal{L}_{DPC}$ by $\mathcal{L}_r$\;
        }
        \Op{}{
              $\mathcal{L}_{Rep} \leftarrow \mathcal{L}_{MCC}+\mathcal{L}_{DPC}$\;
              Update $\mathcal{F}_e$, $\mathcal{F}_c$ and $\mathcal{F}_a$ by $\mathcal{L}_{Rep}$\;
        }
     } 
\end{algorithm}

\textbf{1) Context-aware Semantic (CS) decoder:} To train this decoder, we propose a Multi-patch Context-aware Contrastive (MCC) loss. During training, a randomized image pair $(I_1, I_2|\mathcal{L}_r)$ is sampled from the RPI database. Then, we generate four patches that contain distinct spatial contexts by randomly cropping two patches from each image, as shown in Fig. \ref{fig_genrep_stage}. Each patch is cropped at a random location with a random size $S\times S\times S$, where $S\sim U(\lfloor r_L \cdot S_0\rfloor,\lfloor r_H \cdot S_0\rfloor)$. Here $r_L$ and $r_H$ is the size ratio. These size parameters are set within an appropriate range to ensure that these cropped patches necessarily overlap (see Section \ref{subsec_implementation_details} for detailed parameter values). All patches are subsequently resized to $S_0\times S_0\times S_0$ and augmented via random Gaussian noise. We also generate another two patches that contain a global spatial context by resizing the two original images $I_1$ and $I_2$ to $S_0\times S_0\times S_0$. 

Using these six patches, the CS decoder is trained with an inductive bias that voxels from the same semantic space having similar representations, regardless of appearance and contextual information. We first randomly sample $N_{anc}$ anchor points within the overlapping region of the generated patches. For each patch, the points correspond to the sampled anchor points in the original image are obtained, resulting in $N_{pat} \cdot N_{anc}$ anchor points $\{p_i\}_{i=1}^{N_{pat} \cdot N_{anc}}$, where $N_{pat}$ is the number of patches (here we have $N_{pat}=6$) and $i$ denotes the index of the anchor point ($1\leq i\leq N_{pat} \cdot N_{anc}$). Additionally, we randomly select $N_{can}$ candidate points from all the regions in each patch. For each anchor point $p_i$, we identify $N_{pos,i}$ positive points $\{p_{i,j}^+\}_{j=1}^{N_{pos,i}}$ and $N_{neg,i}$ negative points $\{p_{i,k}^-\}_{k=1}^{N_{neg,i}}$ based on the label map $\mathcal{L}_r$ from all the other anchor points and the candidate points, where $j$ and $k$ are respectively the indexes of the positive point and the negative point, with $1\leq j\leq N_{pos,i}$ and $1\leq k\leq N_{neg,i}$. We define that points from the same labels are positive samples, while points from different labels are negative samples. Let $f_i^c$, $f_{i,j}^{c+}$, and $f_{i,k}^{c-}$ denote respectively the CS embedding of anchor point $p_i$, positive point $p_{i,j}^+$, and negative point $p_{i,k}^-$. The MCC loss $\mathcal{L}_{MCC}$ is formulated as:
\begin{equation}
\begin{aligned}
&\mathcal{L}_{MCC} = -\frac{1}{N_{pat} \cdot N_{anc}} \sum_{i=1}^{N_{pat} \cdot N_{anc}} \bigg ( \frac{1}{N_{pos,i}} \sum_{j=1}^{N_{pos,i}}\\
&  \mathrm{log} \bigg ( \frac{\mathrm{exp}(f_i^c\cdot f_{i,j}^{c+} / \tau_{c})}{\mathrm{exp}(f_i^c\cdot f_{i,j}^{c+} / \tau_{c}) + \sum_{k=1}^{N_{neg,i}} \mathrm{exp}(f_i^c\cdot f_{i,k}^{c-} / \tau_{c})} \bigg ) \bigg )
\end{aligned}
\label{eq_voxel_contrast}
\end{equation}
where $\tau_{c}$ is the temperature parameter. This loss extends the multi-positive supervised contrastive learning framework \citep{khosla2020supervised} by simultaneously enforcing feature affinity among multi-patch positive samples with different spatial contexts while maintaining divergence from negative samples, thereby promoting robust domain-invariant context-aware representations.

\textbf{2) Appearance-aware Semantic (AS) decoder:} This decoder utilizes the Dual-view Prototypical Contrastive (DPC) loss $\mathcal{L}_{DPC}$ \citep{khosla2020supervised}. Specifically, a randomized image pair $(I_1, I_2|\mathcal{L}_r)$ is randomly sampled from the RPI database. Then, label supervision is leveraged for the two views $I_1$ and $I_2$ from the randomized label map $\mathcal{L}_r$ with $L_r$ labels:
\begin{equation}
\begin{aligned}
\mathcal{L}_{DPC} &= - \sum_{l=1}^{L_r} \frac{1}{N_l^v} \sum_{n=1}^{N_l^v} \\
&\mathrm{log} \bigg ( \frac{\mathrm{exp}(f_{l}^a\cdot f_{l,n}^{a} / \tau_a)}{\sum_{l'=1}^{L_r} \sum_{n'=1}^{N_{l'}^v} \mathrm{exp}(f_{l}^a\cdot f_{l',n'}^{a} / \tau_a)} \bigg )
\label{eq_procon}
\end{aligned}
\end{equation}
Here, $f_{l,n}^{a}$ denotes the AS embedding of voxel $n$ with label $l$. Similarly, $f_{l',n'}^{a}$ denotes the AS embedding of voxel $n'$ with label $l'$. $N_l^v$ and $N_{l'}^v$ represent the voxel counts for label $l$ and label $l'$ in the two views, respectively. $f_{l}^a=\frac{1}{N_l^v}\sum_{n=1}^{N_l^v} f_{l,n}^{a}$ serves as the prototype for label $l$. $\tau_a$ is the temperature parameter. This loss establishes a domain-invariant semantic space by minimizing intra-class variations while maximizing inter-class discrepancies.

The CS decoder and the AS decoder are optimized jointly as shown in Algorithm \ref{alg_genrep}. The total loss $\mathcal{L}_{Rep}$ for organ-agnostic semantic representation learning is a combination of the MCC loss $\mathcal{L}_{MCC}$ and the DPC loss $\mathcal{L}_{DPC}$.

\subsubsection{Inference}
After pretraining, for an input image, GSRNet generates the AS embedding $f_a$ from the AS decoder and the CS embedding $f_c$ from the CS decoder. The generalizable semantic embedding $f_s$ is then obtained by concatenating $f_a$ and $f_c$.

\begin{figure}[tbp]
\centering%% For centre alignment of image.
\includegraphics[width=1.0\textwidth]{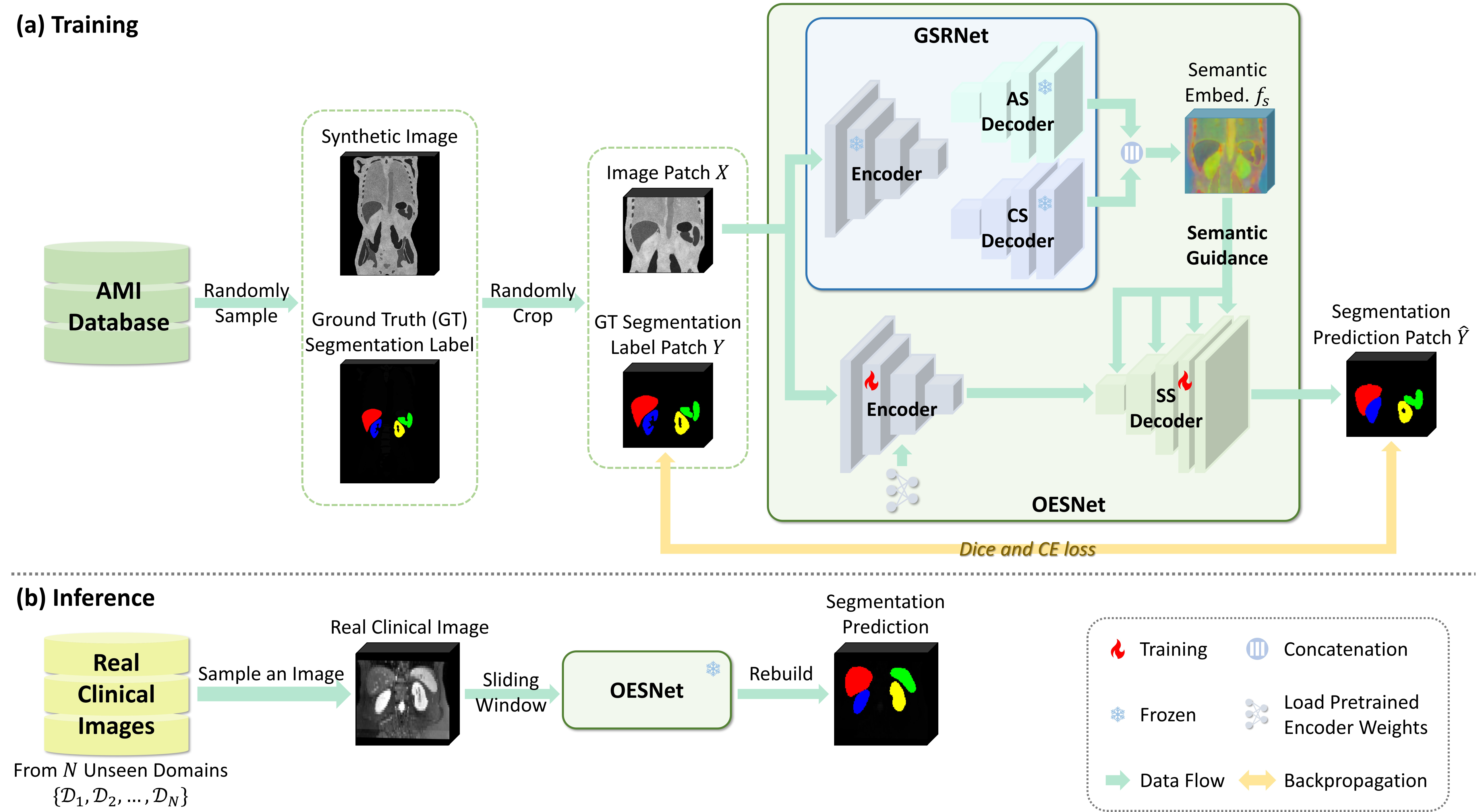}
\caption{Framework for the abdominal-organ-specific segmentation learning stage. During training, a synthetic image with its GT segmentation label is randomly sampled from the AMI database and then randomly cropped to get an image patch with its GT segmentation label patch. The synthetic image patch is fed into an Organ-specific Encoder-decoder Segmentation Network (OESNet), which employs the pretrained weights to initialize the encoder, followed by a Semantics-guided Segmentation (SS) decoder. In the inference stage, OESNet can segment on real images from multiple domains without finetuning. We reduce the dimension of the embeddings to 3 using PCA and visualize them as RGB images. Embed.: embedding. CE: cross entropy.}
\label{fig_genseg_stage}
\end{figure}

\begin{figure}[tbp]
\centering%% For centre alignment of image.
\includegraphics[width=1.0\textwidth]{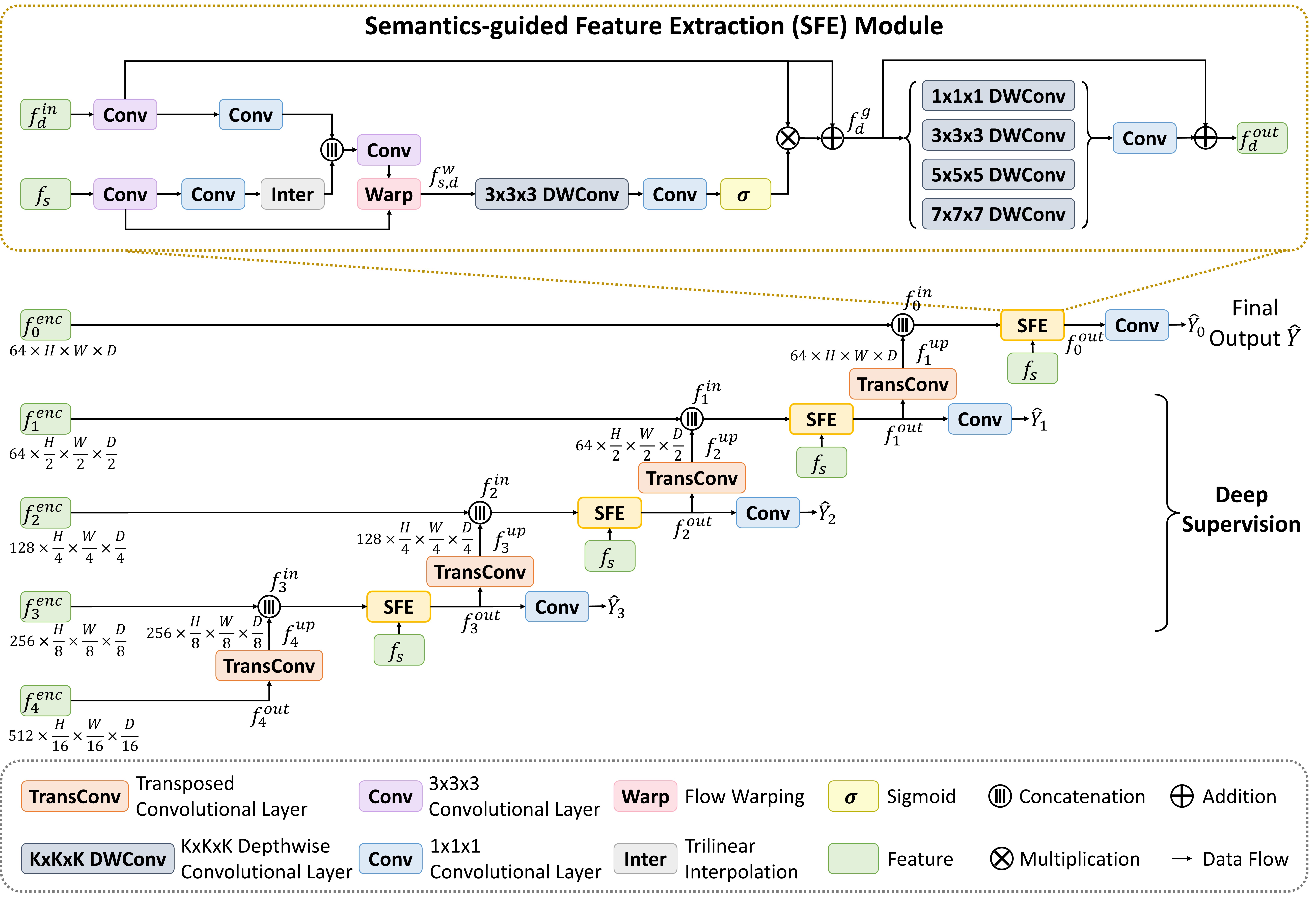}
\caption{Architecture of the SS decoder in OESNet. The SS decoder is integrated with hierarchical SFE modules. Each SFE module takes the segmentation feature of the current layer $f_d^{in}$ and the generalizable semantic embedding $f_s$ from the frozen GSRNet as inputs. Within the SFE module, $f_s$ is first aligned with $f_d^{in}$ and then generates attention map to enhance the segmentation feature. After that, the feature undergoes multi-kernel processing and finally the output feature $f_d^{out}$ is obtained. }
\label{fig_network}
\end{figure}

\subsection{Abdominal-organ-specific segmentation learning}

\subsubsection{Segmentation architecture overview}
This stage employs the pretrained weights of GSRNet to initialize an Organ-specific Encoder-decoder Segmentation Network (OESNet), which are then trained using the AMI database to learn domain-invariant segmentation features (Fig. \ref{fig_genseg_stage}). During training, a GT segmentation label is first sampled from the AMI database, and then a corresponding synthetic image is randomly selected. The training process then operates on a randomly cropped patch  $X\in \mathbb{R}^{H\times W\times D}$ from the synthetic image, with the supervision provided by the spatially aligned GT segmentation label patch $Y \in \mathbb{R}^{H\times W\times D}$. Here $H$, $W$, and $D$ denote respectively the height, width, and depth of the patches.

A schematic illustration of the encoder-decoder architecture of OESNet is depicted in Fig. \ref{fig_genseg_stage}. To achieve generalizable segmentation, we leverage rich semantic priors from the well-trained GSRNet. Specifically, the encoder utilizes the same CNN encoder in GSRNet and is initialized using the pretrained encoder weights. The decoder implements our proposed Semantics-guided Segmentation (SS) mechanism, which incorporates semantic priors from the frozen GSRNet through parallel processing: while the input $X\in \mathbb{R}^{H\times W\times D}$ flows through the main encoder-decoder segmentation network, it is simultaneously processed by the frozen GSRNet to extract the generalizable semantic embedding $f_s \in \mathbb{R}^{C_s \times \frac{H}{2}\times \frac{W}{2}\times \frac{D}{2}}$. Here $C_s$ denotes the feature dimension of $f_s$.

Such semantic embedding aids in identifying salient regions of target organs by providing generalizable cues with rich semantic information acquired during pretraining. Therefore, we introduce the Semantics-guided Feature Extraction (SFE) module, which leverages this generalizable semantic embedding to recalibrate segmentation feature extraction via an attention mechanism. Details about the SFE module will be presented below. As illustrated in Fig. \ref{fig_network}, by integrating four SFE modules hierarchically across four resolution levels ($d\in [0,3]$), the SS decoder receives multi-scale feature extraction guidance from $f_s$. Specifically, at each level $d$, the output feature $f_{d+1}^{out} \in \mathbb{R}^{C_{d+1} \times \frac{H}{2^{d+1}}\times \frac{W}{2^{d+1}}\times \frac{D}{2^{d+1}}}$ from the deeper layer $d+1$ is upsampled via transposed convolution to get $f_{d+1}^{up} \in \mathbb{R}^{C_{d} \times \frac{H}{2^{d}}\times \frac{W}{2^{d}}\times \frac{D}{2^{d}}}$, which is then concatenated with the corresponding encoder feature $f_{d}^{enc}\in \mathbb{R}^{C_d \times \frac{H}{2^d}\times \frac{W}{2^d}\times \frac{D}{2^d}}$ by skip connection. Here $C_d$ denotes the feature dimension at level $d$. The combined feature $f_{d}^{in} \in \mathbb{R}^{2C_d \times \frac{H}{2^d}\times \frac{W}{2^d}\times \frac{D}{2^d}}$ undergoes SFE refinement guided by $f_s$, yielding the output feature $f_{d}^{out}\in \mathbb{R}^{C_d \times \frac{H}{2^d}\times \frac{W}{2^d}\times \frac{D}{2^d}}$ from layer $d$, which produces a segmentation prediction $\hat{Y}_{d}\in \mathbb{R}^{\frac{H}{2^d}\times \frac{W}{2^d}\times \frac{D}{2^d}}$ via an $1\times1\times1$ convolution. 

\subsubsection{Semantics-guided Feature Extraction (SFE) module}
A schematic illustration of the overall architecture of the SFE module is shown in Fig. \ref{fig_network}, which aims to enhance the domain-invariant segmentation feature extraction under the guidance of the generalizable semantic embedding. In this module, the generalizable semantic embedding is first aligned with the segmentation feature, serving as a prerequisite for precise feature extraction guidance. Such alignment ensures that the semantic embedding generates an attention map spatially congruent with the segmentation feature, thereby enabling element-wise multiplication between the attention weights and the segmentation feature to accurately highlight important segmentation feature regions. Initially, the semantic embedding $f_s \in \mathbb{R}^{C_s \times \frac{H}{2}\times \frac{W}{2}\times \frac{D}{2}}$ lacks spatial alignment with the segmentation feature $f_{d}^{in} \in \mathbb{R}^{2C_d \times \frac{H}{2^d}\times \frac{W}{2^d}\times \frac{D}{2^d}}$ due to cross-scale feature propagation via downsampling and upsampling operations \citep{li2024sfnet}. Inspired from the flow alignment method \citep{xie2026biasnet}, we explicitly learn the flow field between $f_s$ and $f_{d}^{in}$ rather than relying on ineffective feature propagation by simple interpolation. Concretely, both features first undergo channel compression to $C_d$ via separate $3\times 3\times 3$ convolutional layers. Then, the two features are projected into a flow estimation space using separate $1\times 1\times 1$ convolutions, followed by trilinear interpolation of $f_s$ to match the spatial dimensions of $f_{d}^{in}$. These features are subsequently concatenated and processed through a $3\times 3\times 3$ convolutional layer to predict a flow field $\bigtriangleup_{d}\in \mathbb{R}^{3 \times \frac{H}{2^d}\times \frac{W}{2^d}\times \frac{D}{2^d}}$. The warped semantic embedding $f_{s,d}^w\in \mathbb{R}^{C_d \times \frac{H}{2^d}\times \frac{W}{2^d}\times \frac{D}{2^d}}$ is then obtained using the differentiable sampling mechanism \citep{jaderberg2015spatial}:
\begin{equation}
f_{s,d}^w = f_s \circ (\Omega_d + \bigtriangleup_{d})
\label{eq_warp}
\end{equation}
where $\Omega_d\in \mathbb{R}^{3 \times \frac{H}{2^d}\times \frac{W}{2^d}\times \frac{D}{2^d}}$ represents the spatial grid coordinates and ``$\circ$'' denotes the warping operation. This learned alignment  establishes the precise feature correspondence essential for subsequent guidance.

Then, we aim to enhance features relevant to segmentation targets while suppressing irrelevant signals, guided by the semantic priors. For this purpose, an attention map is generated from the aligned feature $f_{s,d}^w$, which assigns high response values to target organ regions by parsing the semantic cues. Specifically, a $3\times 3\times 3$ depthwise convolution followed by a $1\times 1\times 1$ convolution produces attention weights, which are activated through a sigmoid function ``$\sigma(\cdot)$''. The resulting attention map modulates the original segmentation features via element-wise multiplication:
\begin{equation}
f_{d}^{g} = f_{d}^{in} + \sigma(\mathrm{Conv} (f_{s,d}^w)) \otimes f_{d}^{in}
\label{eq_gfe}
\end{equation}
where $f_d^g\in \mathbb{R}^{C_d \times \frac{H}{2^d}\times \frac{W}{2^d}\times \frac{D}{2^d}}$ is the refined segmentation feature, ``$\mathrm{Conv}(\cdot)$'' represents the convolutional layers, and ``$\otimes$'' denotes element-wise multiplication. 

Finally, complementary segmentation information is captured across varying receptive fields using parallel multi-kernel processing, as illustrated in Fig. \ref{fig_network}. Specifically, four depthwise convolutional branches with kernel sizes $\{1,3,5,7\}$ each handle one quarter of the feature channels. Additionally,  a residual connection is incorporated to ensure stable gradient flow during backpropagation, preserving critical feature information while enabling effective multi-kernel feature aggregation. 

\subsubsection{Segmentation loss function}
The training objective combines Dice loss and Cross Entropy (CE) loss applied at all four decoder layers ($d\in [0,3]$) with deep supervision. For level $d>0$, we downsample the GT segmentation label patch $Y \in \mathbb{R}^{H\times W\times D}$ to $Y_{d}\in \mathbb{R}^{\frac{H}{2^d}\times \frac{W}{2^d}\times \frac{D}{2^d}}$, which supervises the segmentation prediction $\hat{Y}_{d}\in \mathbb{R}^{\frac{H}{2^d}\times \frac{W}{2^d}\times \frac{D}{2^d}}$. The total segmentation loss function $\mathcal{L}_{Seg}$ is formulated as:
\begin{equation}
\mathcal{L}_{Seg} = \sum_{d=0}^3 w_d (-{\rm DC}(\hat{Y}_{d}, Y_{d}) + {\rm CE}(\hat{Y}_{d}, Y_{d}))
\label{eq_seg}
\end{equation}
where $w_d=\frac{1 / 2^d}{\sum_{d=0}^3 1 / 2^d}$ is the loss weight for level $d$. ``${\rm DC}(\cdot,\cdot)$'' and ``${\rm CE}(\cdot,\cdot)$'' denote the Dice loss function and the CE loss function, respectively. 

\subsubsection{Inference}
\label{subsec_inference}
At inference stage, OESNet can segment target organs on real clinical images across $N$ unseen image domains $\{\mathcal{D}_1,\mathcal{D}_2,\cdots,\mathcal{D}_{N}\}$ without retraining or finetuning. For a real clinical image sampled from an unseen domain, a sliding window approach with overlap ratio of 70\% is adopted to get image patches of size $H\times W \times D$. The sampled image patches are then fed to the trained segmentation network, which uses the highest-resolution prediction $\hat{Y}_0$ as the final segmentation output $\hat{Y}$. We rebuild the corresponding predicted patches to obtain the overall segmentation prediction.

\section{Experiments and results}

This section first describes the experimental setup, then reports segmentation performance across multiple target domains including CT, MRI and PET modalities, and proceeds to validate the effectiveness of the proposed method through comprehensive ablation studies. Finally, the generalizability of our method to different anatomical structures is showcased via extension to the cardiac segmentation task.

\subsection{Datasets}
To assess segmentation generalization capability, we evaluated on seven publicly available abdominal datasets across CT, MRI and PET modalities. Among them, four datasets (two CT datasets and two MR datasets), referred to as the core evaluation datasets, were used for comprehensive evaluation, including comparisons among zero-shot methods, among few-shot methods, and against reference baselines (lower bound and upper bound), as well as ablation analysis. The remaining three datasets (one CT dataset, one MR dataset and one PET dataset), referred to as the generalization evaluation datasets, were used solely to assess zero-shot generalization performance.

The four core evaluation datasets are as follows:
\begin{itemize}
\item \textbf{BTCV:} From the Multi-Atlas Labeling \textbf{B}eyond \textbf{T}he \textbf{C}ranial \textbf{V}ault workshop held at MICCAI 2015 \citep{landman2015miccai}, this dataset contains 30 abdominal CT volumes with 13 annotated organs. For training comparison models that require images in the same domain (dataset-specific methods), we randomly split the data into 15 training/validation and 15 held-out test cases.
\item \textbf{L2RAb-MR:} From the \textbf{L}earn\textbf{2R}eg \textbf{Ab}domen MRI-CT challenge \citep{hering2022learn2reg}, this collection combines 40 T2-SPIR MR images from the Combined Healthy Abdominal Organ Segmentation (CHAOS) challenge \citep{CHAOS2021} and 8 multi-phase MR images from The Cancer Imaging Archive (TCIA) project \citep{clark2013cancer}, each of which has 4 annotated organs. For quality control, we excluded 3 cases with erroneous annotations in automatically segmented CHAOS test subset provided by Learn2Reg, leaving 45 valid volumes. These were randomly divided into 23 training/validation cases for dataset-specific methods and 22 held-out test samples.
\item \textbf{AMOS-CT:} This dataset is from the Multi-Modality \textbf{A}bdominal \textbf{M}ulti-\textbf{O}rgan \textbf{S}egmentation Challenge \citep{ji2022amos} held at MICCAI 2022, which provides 500 CT scans collected from multi-center, multi-vendor, and multi-disease patients, each with 15 annotated organs. The data were officially partitioned into 200 training, 100 validation, and 200 private test cases. We adopted the original training set (200) as the training and validation data for dataset-specific methods and the official validation set (100) as the test data.
\item \textbf{AMOS-MR:} The MR counterpart of AMOS-CT \citep{ji2022amos} contains 100 multi-phase scans (15 annotated organs) with official splits: 40 training, 20 validation, and 40 private test volumes. We utilized the 40 training scans for training and validation of dataset-specific methods and the original 20 validation cases for testing.
\end{itemize}

The three generalization evaluation datasets are as follows:
\begin{itemize}
\item \textbf{WORD:} The \textbf{W}hole abdominal \textbf{OR}gan \textbf{D}ataset \citep{luo2022word} consists of 150 CT volumes, each with 16 annotated organs. All the cases were used for validating the generalization capability of models.
\item \textbf{MRISegmentator:} An abdominal MR dataset \citep{zhuang2024mrisegmentator,zhuang2025mrisegmenter} comprises of axial precontrast T1, arterial, venous, and delayed phases for each patient. Each scan contains 62 annotated abdominal structures. We used the official 240 test cases to evaluate the generalization capability of models.
\item \textbf{AutoPET-Organ:} This abdominal PET dataset is extended from the AutoPET dataset \citep{gatidis2022whole} by \cite{zhang2025seganypet}, who annotated 12 organs on a small subset with 100 cases from the original 1,014 3D whole-body PET images. All the annotated cases were employed for testing.
\end{itemize}

For all datasets, to generate the GT segmentation labels for evaluation, we extracted the labels corresponding to our target abdominal organs (the liver, the spleen, the right kidney, and the left kidney) from the annotations, renumbered them consecutively, and assigned the background label to non-target structures.

\subsection{Comparison methods}
We compared ReSeg against both zero-shot and few-shot approaches:
\begin{itemize}
\item \textbf{SAM-based zero-shot methods:} These methods leverage foundation models like SAM \citep{kirillov2023segment} for data-efficient 2D medical image segmentation. We benchmarked two SOTA SAM-based methods: 1) \textbf{MedCLIP-SAM} \citep{koleilat2024medclip}, which implements text-prompted segmentation in a strict zero-shot setting; and 2) \textbf{ZeroShotSAM} \citep{shaharabany2024zero}, which performs minimally-guided zero-shot segmentation using $M$ foreground points per 2D slice as prompts. 
\item \textbf{Synthetic data-based zero-shot method:} We compared against \textbf{SynthSeg} \citep{billot2023synthseg}, a method that demonstrates domain generalization for brain scans by training exclusively on synthetic data generated via a domain randomization strategy. In our abdominal organ segmentation experiment, we also employed the segmentation masks in TotalSegmentator CT dataset \citep{wasserthal2023totalsegmentator} to train \textbf{SynthSeg} \citep{billot2023synthseg} for fair comparison.
\item \textbf{Meta-learning-based few-shot methods:} These methods segment unseen classes by capturing similarities between $K$ labeled support samples and test images. We compared three SOTA prototype-based few-shot approaches: \textbf{SSL-ALPNet} \citep{ouyang2022self}, \textbf{DSPNet} \citep{tang2025few}, and \textbf{DIFD} \citep{cheng2025dual}. All models are trained on in-domain real images using self-supervised learning with superpixel pseudo-labels. For training, we followed the less strict experimental setting in \cite{roy2020squeeze}, where test classes may appear in training data but remain unlabeled. In addition, we compared \textbf{UniverSeg} \citep{butoi2023universeg}, a typical few-shot foundation model trained on a large scale of medical image datasets.
\item \textbf{Finetuning-based few-shot methods:} These methods implement few-shot segmentation by first pretraining a strong biomedical foundation model and then finetune it with $K$ annotated volumes. To achieve strong performance, extensive augmentation and full-network finetuning are employed. We compared two SOTA representation learning methods: 1) \textbf{VoCo} \citep{wu2024voco},  a simple-yet-effective volume contrast framework leveraging geometric relations between different organs for pretraining, which requires real training images; and 2) \textbf{Anatomix} \citep{dey2025learning}, a representation learning method using a domain randomization strategy to synthesize highly variable training samples, enabling networks to acquire general-purpose biomedical volume representations during pretraining. 
\end{itemize}

To quantify the performance of different competing methods, we trained two additional models as the reference baselines for each core evaluation dataset. These two models were developed using nnUNet \citep{isensee2021nnu}, a SOTA supervised method that automatically configures preprocessing, network architecture, and training. The first model, referred to as the upper bound model, was trained on the training data of one dataset and then tested on the testing data of the same dataset. In contrast, the second model, referred to as the lower bound model, was trained and tested on datasets from different modalities. Specifically, the lower bound model for the CT datasets was trained on the L2RAb-MR dataset, while the lower bound model for the MR datasets was trained on the BTCV CT dataset.

\subsection{Implementation details}
\label{subsec_implementation_details}
The proposed ReSeg framework was implemented using PyTorch v2.1. For the pretraining stage, we configured each decoder with an embedding dimension of 128. During contrastive learning, the AS decoder processed entire paired synthetic volumes of size $128\times128\times128$ with batch size 1, while the CS decoder utilized six overlapping patches of size $96\times96\times96$ with batch size 2 ($S_0=96$). We empirically set $r_L=0.8$, $r_H=1.2$, $N_{anc}=256$, $N_{can}=512$, $\tau_c=0.33$ and $\tau_a=0.1$. We used Stochastic Gradient Descent (SGD) with a momentum of 0.9 and a weight decay of $1.0\times10^{-4}$, employing a learning rate of 0.02 with warmup and step decay scheduling. This pretraining phase lasted for a total of 45,000 iterations.

For the segmentation learning stage, we used input patches of size $128\times128\times128$ with $3.0\times3.0\times3.0~\mathrm{mm}^3$ spacing. Training employed a batch size of 2, and OESNet was optimized using SGD with the same momentum and weight decay parameters as in the pretraining stage, but with a reduced learning rate of 0.01 that similarly incorporated warmup and step decay. This stage required 80,000 training iterations.

For all comparative methods, we implemented them using official codebases. ZeroShotSAM~\citep{shaharabany2024zero} received minimal guidance through $M=1$ foreground point per 2D slice. Few-shot methods used identical support images ($K=1$) randomly selected from the training subsets.

\subsection{Evaluation protocol}
To evaluate the capability of the segmentation models, we employed the Dice Similarity Coefficient (DSC) \citep{dice1945measures}, which is a standard metric used to evaluate 3D medical image segmentation \citep{isensee2021nnu}. DSC quantifies the spatial overlap between segmentation predictions and GT segmentation labels. Following the evaluation protocol introduced in \citep{isensee2021nnu}, we first computed the DSC score for each target organ per case, then averaged the scores across all cases to get the organ-wise DSC scores, and finally derived the average DSC by averaging across all target organs.

The DSC scores were evaluated in the original space of the testing 3D volumes. For 2D benchmarks, we processed volumetric data through vertical axial slice decomposition. For 3D models, we directly processed the entire 3D images to predict 3D segmentation masks.

\begin{table}[tbp]%% placement specifier
\centering%% For centre alignment of tabular.
\caption{Quantitative results for the comparison with the SOTA methods on both BTCV and AMOS-CT datasets. The best results are highlighted in bold, and the second-best results are marked with underlining. The values in parentheses indicate the standard deviation of the DSC scores. R-Kid: right kidney. L-Kid: left kidney. MGZS: minimally-guided zero-shot. ZS: zero-shot. MLFS: meta-learning-based few-shot. FTFS: finetuning-based few-shot. }
\adjustbox{width=\textwidth}{
\begin{tabular}{lcc|ccccc|ccccc}
\toprule
\multirow{2}{*}{Method} & Method & \multirow{2}{*}{2D/3D} & \multicolumn{5}{c}{DSC (\%) Results on the BTCV dataset} & \multicolumn{5}{|c}{DSC (\%) Results on the AMOS-CT dataset} \\
\cmidrule(lr){4-8} 
\cmidrule(lr){9-13} 
 &  Type &  & Liver & Spleen & R-Kid & L-Kid & Average & Liver & Spleen & R-Kid & L-Kid & Average \\
\midrule
\multicolumn{13}{l}{\textbf{\textit{Reference Baselines}}} \\
Lower Bound  & - &  3D &  34.92 (9.24) & 69.73 (27.30) & 73.61 (14.00) & 71.82 (21.78) & 62.52 (14.75) & 23.90 (8.70) & 32.92 (33.01) & 47.02 (33.81) & 42.94 (37.95) & 36.70 (24.67) \\
Upper Bound & - & 3D & 94.16 (9.28) & 95.77 (3.53) & 93.97 (2.64) & 93.82 (2.50) & 94.43 (2.87) & 97.67 (1.09) & 96.89 (3.54) & 96.68 (1.38) & 96.64 (1.83) & 96.97 (1.27) \\
\midrule
\multicolumn{13}{l}{\textbf{\textit{Zero-shot Methods}}} \\
\makecell[l]{ZeroShotSAM \\ \citep{shaharabany2024zero}}  & MGZS & 2D &  80.37 (11.45) & 79.00 (17.05) & \textbf{91.24 (2.27)} & \textbf{91.26 (2.78)} & 85.47 (6.05) & 68.12 (20.02) & \underline{73.53 (26.72)} & \underline{84.89 (18.49)} & \underline{87.10 (15.36)} & \underline{78.41 (16.80)} \\
\makecell[l]{MedCLIP-SAM \\ \citep{koleilat2024medclip}}  & ZS & 2D & 31.16 (17.12) & 1.10 (2.88) & 1.97 (4.49) & 0.68 (2.53) & 8.73 (4.46) & 45.51 (13.57) & 4.54 (8.62) & 3.60 (5.91) & 2.45 (6.17) & 14.02 (4.72)  \\
\makecell[l]{SynthSeg \\ \citep{billot2023synthseg}}  & ZS  & 3D &  34.19 (39.12) & 43.15 (45.09) & 76.12 (23.98) & 73.82 (30.69) & 56.82 (27.45)  & 34.31 (40.13) & 33.25 (40.67) & 56.80 (40.47) & 65.69 (37.90) & 47.51 (33.59) \\
\makecell[l]{\textbf{ReSeg} \\ \textbf{(Ours)}}  & ZS & 3D & 90.13 (6.49) & \textbf{90.26 (4.05)}& 84.75 (9.21) & \underline{87.43 (4.08)} & \textbf{88.14 (3.91)} & \textbf{90.76 (4.74)} & \textbf{88.80 (5.80)} & \textbf{86.80 (4.25)} & \textbf{88.21 (4.78)} & \textbf{88.64 (3.20)}  \\
\midrule
\multicolumn{13}{l}{\textbf{\textit{Few-shot Methods}}} \\
\makecell[l]{SSL-ALPNet \\ \citep{ouyang2022self}}  & MLFS & 2D & 80.09 (4.80) & 78.29 (6.15) & 77.17 (9.34) & 76.98 (5.71) & 78.13 (4.42) & 68.53 (7.38) & 26.17 (18.34) & 70.87 (11.67) & 72.70 (9.36) & 59.57 (7.56)  \\
\makecell[l]{DSPNet \\ \citep{tang2025few}}  & MLFS & 2D &  78.84 (5.23) & 73.95 (9.36) & 77.95 (7.33) & 76.19 (6.73) & 76.73 (4.08) & 67.90 (7.24) & 38.82 (15.68) & 59.67 (14.25) & 75.35 (7.87) & 60.44 (7.25)   \\
\makecell[l]{DIFD \\ \citep{cheng2025dual}}  & MLFS & 2D & 79.48 (6.88) & 83.08 (6.11) & 84.36 (7.24) & 75.24 (13.99) & 80.54 (4.31) & 63.24 (8.91) & 11.20 (14.56) & 60.52 (19.16) & 77.19 (13.67) & 53.03 (8.58) \\
\makecell[l]{UniverSeg \\ \citep{butoi2023universeg}}  & MLFS  &  2D & 76.54 (13.22) & 50.35 (34.15) & 49.75 (27.55) & 57.66 (19.03) & 58.58 (15.84) & 76.06 (10.99) & 16.30 (17.87) & 59.97 (20.71) & 60.12 (19.11) & 53.11 (11.36)  \\
\makecell[l]{VoCo \\ \citep{wu2024voco}}  & FTFS  & 3D & \textbf{92.59 (6.49)} & \underline{83.29 (11.86)} & \underline{85.27 (11.62)} & 81.71 (12.44) & \underline{85.72 (8.64)} &  \underline{86.20 (10.30)} & 66.47 (20.56) & 69.87 (21.31) & 66.65 (16.67) & 72.30 (12.19)   \\
\makecell[l]{Anatomix \\ \citep{dey2025learning}}  & FTFS  & 3D &  \underline{91.20 (5.33)} & 81.66 (8.35) & 75.18 (22.77) & 80.47 (16.28) & 82.13 (9.29) & 83.43 (7.03) & 49.37 (25.08) & 77.74 (11.21) & 68.91 (28.42) & 69.86 (12.02) \\
\bottomrule
\end{tabular}
}
\label{tab_compare_ct}
\end{table}

\begin{figure}[tbp]
\centering%% For centre alignment of image.
\includegraphics[width=1.0\textwidth]{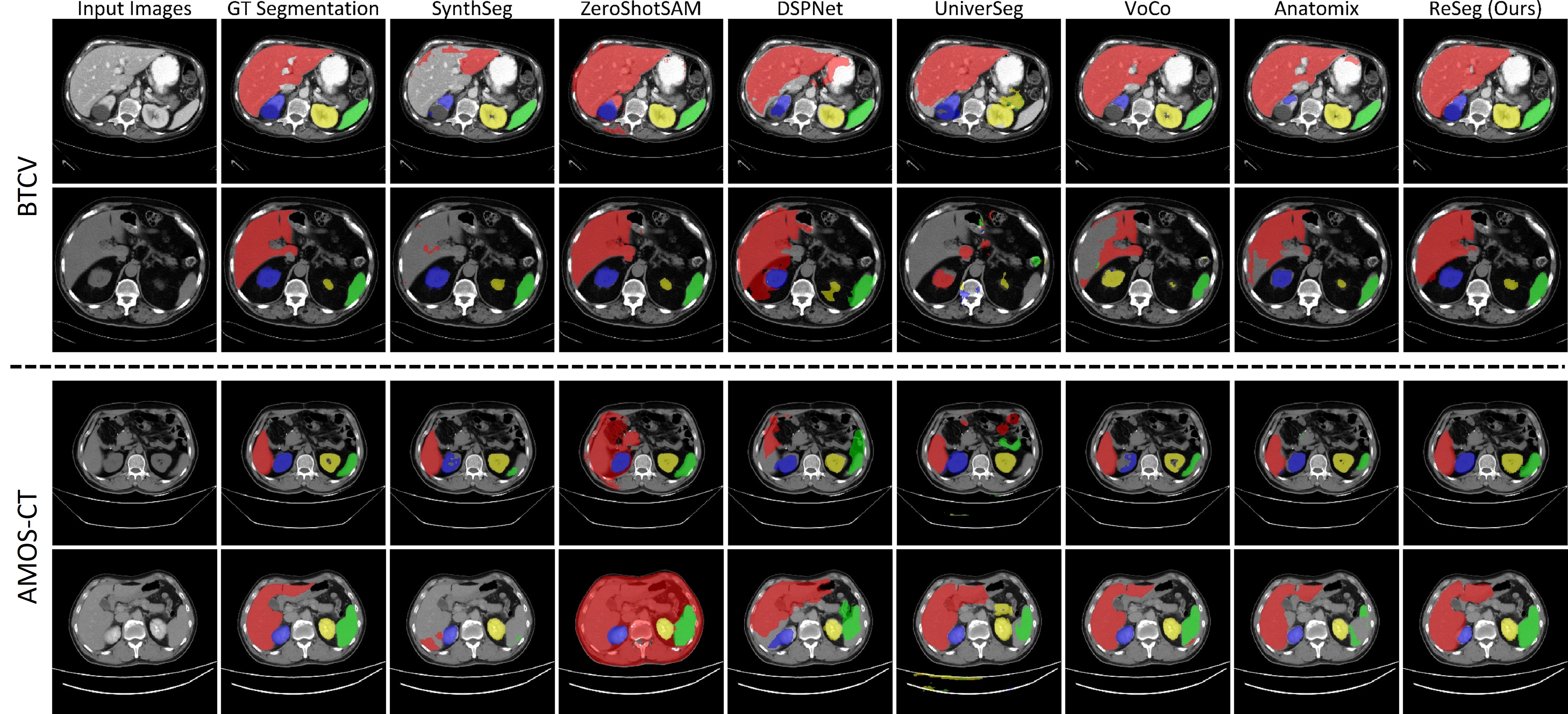}
\caption{Visual results for the comparison with the SOTA methods on both BTCV and AMOS-CT datasets. 3D examples are shown in 2D slices along the transverse plane. Left to right: input images, GT segmentation labels, and segmentation results of different methods including SynthSeg \citep{billot2023synthseg}, ZeroShotSAM \citep{shaharabany2024zero}, DSPNet \citep{tang2025few}, UniverSeg \citep{butoi2023universeg}, VoCo \citep{wu2024voco}, Anatomix \citep{dey2025learning}, and the proposed ReSeg. Colors correspond to the organs as follows: red - the liver, green - the spleen, blue - the right kidney, yellow - the left kidney.}
\label{fig_compare_ct}
\end{figure}

\subsection{Comparison results}
\subsubsection{Results on the core evaluation datasets}
\textbf{1) Quantitative results on the CT datasets:} We first investigated the performance of the reference baselines on both BTCV and AMOS-CT datasets. The results are summarized in Table~\ref{tab_compare_ct}. Specifically, the upper bound model achieved an average DSC of $94.43\%$ on the BTCV dataset, but this dropped sharply to $62.52\%$ under the lower bound setting. Similarly, on the AMOS-CT dataset, the average DSC fell from $96.97\%$ to $36.70\%$. Such a large discrepancy highlights the fundamental challenge of cross-modality generalization. 

We then compared our proposed ReSeg with both zero-shot and few-shot competing methods on these two datasets. As shown in Table~\ref{tab_compare_ct}, among all the zero-shot competing methods, ZeroShotSAM~\citep{shaharabany2024zero} obtained the best results. In particular, ZeroShotSAM~\citep{shaharabany2024zero} reported an average DSC of $85.47\%$ on the BTCV dataset and an average DSC of $78.41\%$ on the AMOS-CT dataset. Among all the few-shot competing methods, VoCo~\citep{wu2024voco} obtained the best results on these two datasets, yielding respectively an average DSC of $85.72\%$ and $72.30\%$. Overall, ReSeg achieved the best performance. Specifically, ReSeg achieved respectively an average DSC of $88.14\%$ and $88.64\%$ on these two datasets. Fig. \ref{fig_compare_ct} provides a visual comparison between our method and the competing methods when evaluated on both BTCV and AMOS-CT datasets. It can be observed that our method consistently delivers the most precise boundary delineation among the zero-shot and few-shot competing approaches, visually confirming its superior zero-shot segmentation capability.

\begin{table}[tbp]%% placement specifier
\centering%% For centre alignment of tabular.
\caption{Quantitative results for the comparison with the SOTA methods on both L2RAb-MR and AMOS-MR datasets. The best results are highlighted in bold, and the second-best results are marked with underlining. The values in parentheses indicate the standard deviation of the DSC scores. R-Kid: right kidney. L-Kid: left kidney. MGZS: minimally-guided zero-shot. ZS: zero-shot. MLFS: meta-learning-based few-shot. FTFS: finetuning-based few-shot. }
\adjustbox{width=\textwidth}{
\begin{tabular}{lcc|ccccc|ccccc}
\toprule
\multirow{2}{*}{Method} & Method & \multirow{2}{*}{2D/3D} & \multicolumn{5}{c}{DSC (\%) Results on the L2RAb-MR dataset} & \multicolumn{5}{|c}{DSC (\%) Results on the AMOS-MR dataset} \\
\cmidrule(lr){4-8} 
\cmidrule(lr){9-13} 
 &  Type &  & Liver & Spleen & R-Kid & L-Kid & Average & Liver & Spleen & R-Kid & L-Kid & Average \\
\midrule
\multicolumn{13}{l}{\textbf{\textit{Reference Baselines}}} \\
Lower Bound  & - &  3D & 7.99 (25.29) & 4.78 (15.12) & 3.85 (17.64) & 7.88 (24.40) & 6.12 (17.88)  & 27.20 (25.02) & 23.47 (33.47) & 38.37 (39.09) & 44.94 (37.75) & 33.49 (24.11) \\
Upper Bound  & - & 3D & 91.88 (13.13) & 87.50 (20.07) & 93.96 (6.17) & 93.16 (5.63) & 91.62 (8.89) & 98.26 (0.93) & 98.14 (0.98) & 97.37 (0.62) & 97.51 (0.46) & 97.82 (0.56) \\
\midrule
\multicolumn{13}{l}{\textbf{\textit{Zero-shot Methods}}}\\
\makecell[l]{ZeroShotSAM \\ \citep{shaharabany2024zero}}  & MGZS & 2D & 55.04 (18.74) & 71.04 (26.17) & 82.85 (16.59) & \underline{85.36 (12.66)} & 73.57 (13.09)  & 73.15 (16.59) & \underline{72.06 (26.80)} & 86.23 (9.90) & \underline{90.51 (4.59)} & 80.49 (11.50) \\
\makecell[l]{MedCLIP-SAM \\ \citep{koleilat2024medclip}}  & ZS & 2D &  10.14 (13.65) & 10.98 (11.78) & 9.68 (7.94) & 14.88 (11.25) & 11.42 (5.37)  & 29.75 (20.35) & 5.28 (7.02) & 7.97 (9.17) & 8.86 (10.32) & 12.96 (5.43)\\
\makecell[l]{SynthSeg \\ \citep{billot2023synthseg}}  & ZS  & 3D & 9.66 (21.31) & 15.46 (27.31) & 35.71 (34.57) & 75.16 (14.46) & 33.99 (20.42)  & 15.24 (21.30) & 55.75 (34.80) & 44.08 (39.79) & 74.58 (24.85) & 47.41 (21.29)  \\
\makecell[l]{\textbf{ReSeg} \\ \textbf{(Ours)}}  & ZS & 3D & \textbf{86.16 (7.35)} & \textbf{86.63 (5.88)} & \underline{86.49 (6.17)} & \textbf{87.30 (5.48)} & \textbf{86.64 (3.73)} & \underline{89.69 (7.09)} & \textbf{90.81 (2.23)} & \textbf{89.81 (1.77)} & \textbf{91.21 (1.61)} & \textbf{90.38 (1.87)} \\
\midrule
\multicolumn{13}{l}{\textbf{\textit{Few-shot Methods}}} \\
\makecell[l]{SSL-ALPNet \\ \citep{ouyang2022self}}  & MLFS & 2D & 79.25 (3.96) & 76.56 (14.17) & 77.86 (11.15) & 79.59 (9.74) & 78.32 (6.40)  & 75.50 (4.36) & 65.89 (12.99) & 73.01 (17.78) & 73.70 (12.18) & 72.03 (8.97)  \\
\makecell[l]{DSPNet \\ \citep{tang2025few}}  & MLFS & 2D &  78.16 (4.45) & 77.31 (13.31) & 77.22 (13.55) & 79.11 (11.41) & 77.95 (6.74) & 75.91 (4.02) & 71.18 (8.89) & 74.39 (11.44) & 74.26 (8.42) & 73.94 (5.17)  \\
\makecell[l]{DIFD \\ \citep{cheng2025dual}}  & MLFS & 2D & 75.40 (6.71) & 74.38 (23.19) & 76.02 (16.46) & 74.00 (20.79) & 74.95 (12.28)  & 56.80 (6.60) & 26.76 (20.02) & 37.26 (29.77) & 46.08 (32.40) & 41.72 (19.13)  \\
\makecell[l]{UniverSeg \\ \citep{butoi2023universeg}}  & MLFS  &  2D &  79.25 (11.13) & 77.76 (20.17) & 60.01 (23.70) & 75.55 (17.42) & 73.14 (12.08)  & 78.93 (6.60) & 36.53 (24.66) & 78.47 (15.39) & 71.91 (18.50) & 66.46 (14.29)  \\
\makecell[l]{VoCo \\ \citep{wu2024voco}}  & FTFS  & 3D &  \underline{80.66 (24.62)} & 69.31 (29.50) & 62.57 (28.35) & 79.83 (11.81) & 73.09 (20.63)  & 88.39 (5.25) & 32.85 (22.71) & 78.17 (25.53) & 88.20 (6.29) & 71.90 (6.83) \\
\makecell[l]{Anatomix \\ \citep{dey2025learning}}  & FTFS  & 3D &  79.16 (17.94) & \underline{78.82 (24.60)} & \textbf{87.57 (17.53)} & 78.14 (19.41) & \underline{80.93 (16.57)}  & \textbf{92.88 (5.11)} & 70.38 (28.93) & \underline{87.18 (14.60)} & 87.53 (9.30) & \underline{84.49 (10.81)} \\
\bottomrule
\end{tabular}
}
\label{tab_compare_mr}
\end{table}

\begin{figure}[tbp]
\centering%% For centre alignment of image.
\includegraphics[width=1.0\textwidth]{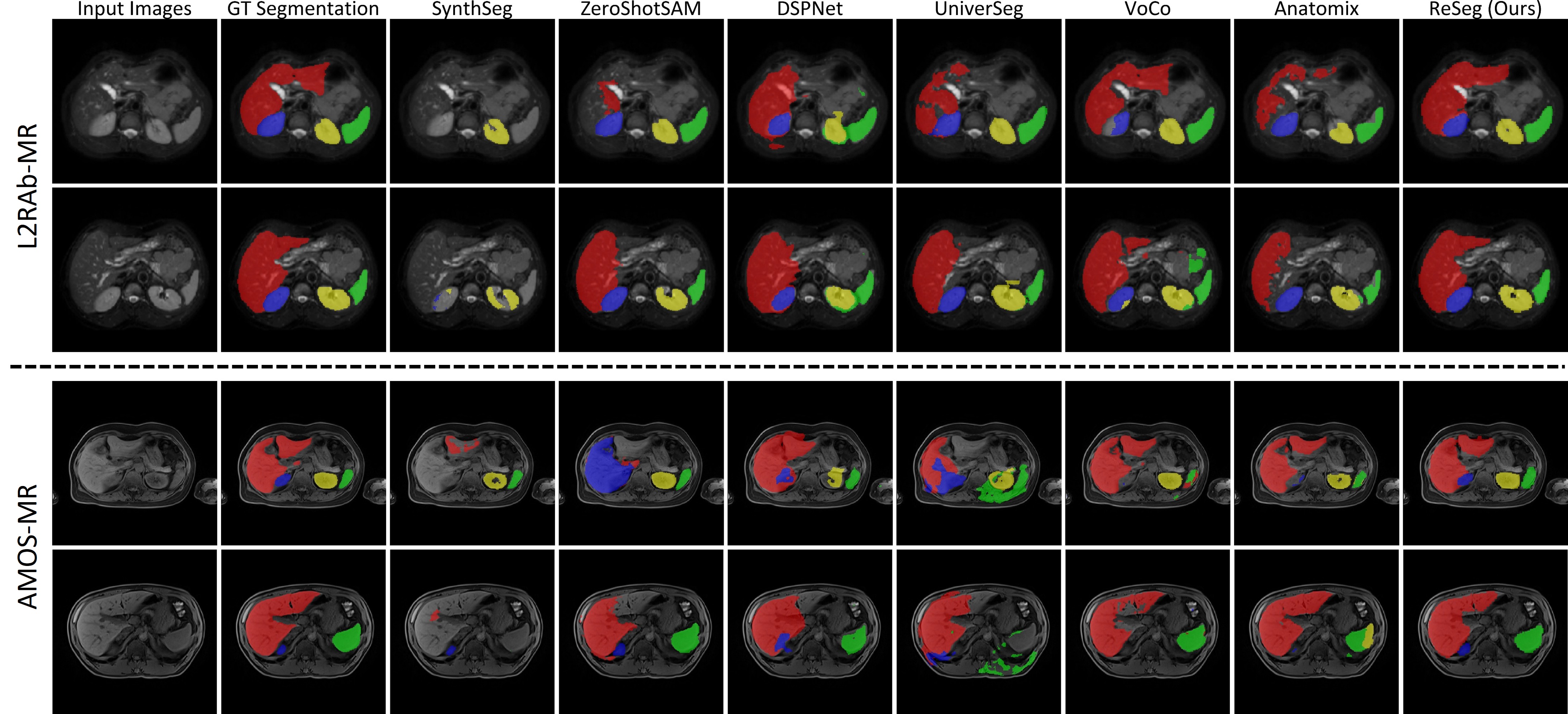}
\caption{Visual results for the comparison with the SOTA methods on both L2RAb-MR and AMOS-MR datasets. 3D examples are shown in 2D slices along the transverse plane. Left to right: input images, GT segmentation labels, and segmentation results of different methods including SynthSeg \citep{billot2023synthseg}, ZeroShotSAM \citep{shaharabany2024zero}, DSPNet \citep{tang2025few}, UniverSeg \citep{butoi2023universeg}, VoCo \citep{wu2024voco}, Anatomix \citep{dey2025learning}, and the proposed ReSeg. Colors correspond to the organs as follows: red - the liver, green - the spleen, blue - the right kidney, yellow - the left kidney.}
\label{fig_compare_mr}
\end{figure}

\textbf{2) Quantitative results on the MR datasets:} We began by investigating the performance of the reference baselines on both L2RAb-MR and AMOS-MR datasets. The results  are presented in Table \ref{tab_compare_mr}.  As reported in this table, the performance of the lower bound model was significantly compromised, yielding respectively an average DSC of $6.12\%$ and $33.49\%$ on these two datasets. In contrast, the upper bound model achieved respectively an average DSC of $91.62\%$ and $97.82\%$ on these two datasets. These results consistently reveal substantial performance degradation when models are tested on data from new domains. 

We further evaluated the performance of the proposed ReSeg and the competing methods on these two datasets. As shown in Table \ref{tab_compare_mr}, ReSeg achieved the best results in terms of average DSC when compared to both zero-shot and few-shot competing methods. Notably, ReSeg attained an average DSC of $86.64\%$ on the L2RAb-MR dataset and an average DSC of $90.38\%$ on the AMOS-MR dataset. In comparison, the best zero-shot competing method (ZeroShotSAM~\citep{shaharabany2024zero}) obtained respectively an average DSC of $73.57\%$ and $80.49\%$ on these two datasets, while the best few-shot competing method (Anatomix~\citep{dey2025learning}) obtained respectively an average DSC of $80.93\%$ and $84.49\%$ on the same two datasets.  Fig. \ref{fig_compare_mr} visually compares the segmentation results achieved by different methods on these two datasets. One can see from this figure that ReSeg consistently produces the most accurate organ boundaries, further substantiating its superior performance.

\begin{figure}[tbp]
\centering%% For centre alignment of image.
\includegraphics[width=1.0\textwidth]{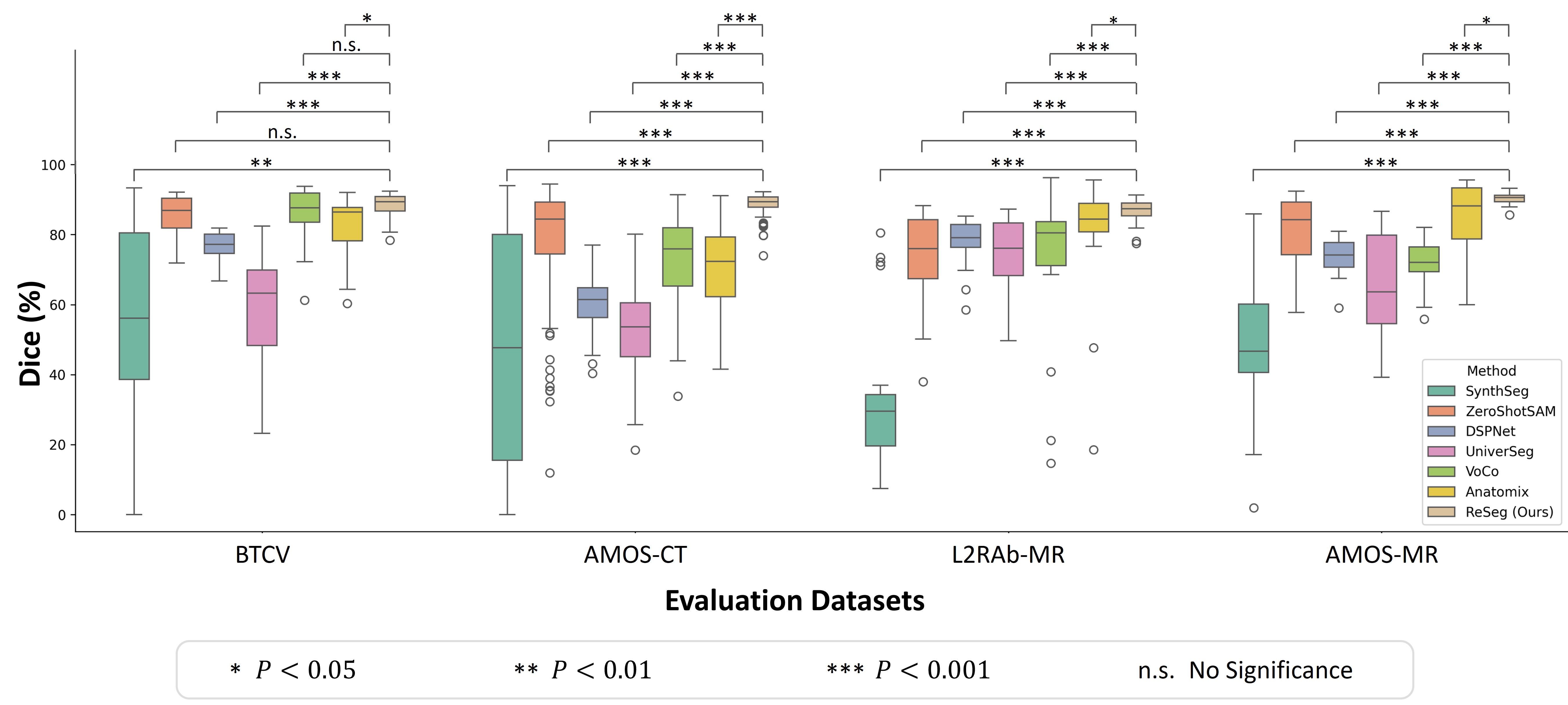}
\caption{Box plots of the DSC scores across the BTCV, the AMOS-CT, the L2RAb-MR, and the AMOS-MR datasets, demonstrating the performance of ReSeg versus SOTA methods. Statistical significances from Wilcoxon signed-rank test \citep{woolson2007wilcoxon} are annotated above the box plots.}
\label{fig_compare_box}
\end{figure}

\textbf{3) Statistical analysis:} We further present a comprehensive statistical comparison of segmentation performance on the BTCV, the AMOS-CT, the L2RAb-MR, and the AMOS-MR datasets through box plot visualization, as illustrated in Fig. \ref{fig_compare_box}. The analysis was conducted by computing the case-wise DSC scores through organ-wise averaging for each method, with the box plots representing the median (central line), the first quartile (lower edge), and the third quartile (upper edge) of these scores. We further conducted paired Wilcoxon signed-rank tests \citep{woolson2007wilcoxon} and chose a significant level of 0.05 in order to compare ReSeg against both zero-shot and few-shot competing methods. The statistical testing results were shown above the corresponding box plots. From this figure, one can see that ReSeg consistently achieves superior median DSC scores with notably tighter distributions, indicating more reliable and precise segmentation performance. Statistical significance test further confirms the performance advantages of our method, particularly on the AMOS-CT, the L2RAb-MR, and the AMOS-MR datasets, where the improvements over the competing approaches are all statistically significant.

\begin{table}[tbp]%% placement specifier
\centering%% For centre alignment of tabular.
\caption{Quantitative results obtained by comparing the generalization capability of different zero-shot segmentation methods when evaluated on the WORD, the MRISegmentator, and the AutoPET-Organ datasets. The best results are highlighted in bold, and the second-best results are marked with underlining. The values in parentheses indicate the standard deviation of the DSC scores. R-Kid: right kidney. L-Kid: left kidney. }
\adjustbox{width=\textwidth}{
\begin{tabular}{l|ccccc}
\toprule
Method & Liver & Spleen & R-Kid & L-Kid & Average  \\
\midrule
\multicolumn{6}{l}{\textbf{\textit{DSC (\%) Results on the WORD dataset}}} \\
ZeroShotSAM \citep{shaharabany2024zero} &  \underline{71.74 (15.71)} & \underline{82.21 (18.43)} & \textbf{91.71 (8.12)} & \textbf{91.35 (8.82)} & \underline{84.25 (9.22)}  \\
MedCLIP-SAM \citep{koleilat2024medclip} & 54.17 (18.53) & 2.79 (5.92) & 5.46 (7.55) & 4.17 (6.91) & 16.65 (5.48) \\
SynthSeg \citep{billot2023synthseg} &  32.42 (37.89) & 29.00 (38.57) & 39.34 (42.29) & 53.12 (42.09) & 38.47 (35.11)  \\
\textbf{ReSeg (Ours)}  &  \textbf{91.50 (4.16)} & \textbf{88.58 (4.90)} & \underline{88.97 (5.49)} & \underline{89.42 (8.12)} & \textbf{89.62 (4.56)} \\
\midrule
\multicolumn{6}{l}{\textbf{\textit{DSC (\%) Results on the MRISegmentator dataset}}} \\
ZeroShotSAM \citep{shaharabany2024zero} & \underline{81.95 (9.88)} & \underline{76.37 (20.89)} & \underline{86.10 (13.59)} & \underline{85.05 (15.18)} & \underline{82.37 (11.09)}   \\
MedCLIP-SAM \citep{koleilat2024medclip} &  39.03 (16.85) & 7.68 (10.14) & 10.97 (14.09) & 13.06 (15.15) & 17.69 (8.06) \\
SynthSeg \citep{billot2023synthseg} & 27.32 (27.75) & 43.86 (39.59) & 50.29 (35.31) & 69.73 (26.72) & 47.80 (25.54)   \\
\textbf{ReSeg (Ours)}  &  \textbf{91.85 (6.67)} & \textbf{89.71 (6.05)} & \textbf{89.77 (3.28)} & \textbf{88.84 (10.13)} & \textbf{90.04 (3.64)} \\
\midrule
\multicolumn{6}{l}{\textbf{\textit{DSC (\%) Results on the AutoPET-Organ dataset}}} \\
ZeroShotSAM \citep{shaharabany2024zero} &  14.19 (12.20) & \underline{7.66 (3.61)} & \underline{22.18 (20.65)} & \underline{22.67 (22.75)} & \underline{16.68 (10.72)}  \\
MedCLIP-SAM \citep{koleilat2024medclip} & \underline{33.94 (27.28)} & 1.39 (3.96) & 13.15 (10.62) & 11.30 (12.61) & 14.95 (8.34) \\
SynthSeg \citep{billot2023synthseg} & 0.00 (0.05) & 0.12 (0.72) & 0.71 (4.96) & 7.76 (19.32) & 2.15 (5.15)   \\
\textbf{ReSeg (Ours)}  & \textbf{82.81 (17.27)} & \textbf{69.06 (27.13)} & \textbf{68.49 (20.37)} & \textbf{69.24 (23.80)} & \textbf{72.40 (18.04)}  \\
\bottomrule
\end{tabular}
}
\label{tab_compare_zeroshot}
\end{table}

\begin{figure}[tbp]
\centering%% For centre alignment of image.
\includegraphics[width=1.0\textwidth]{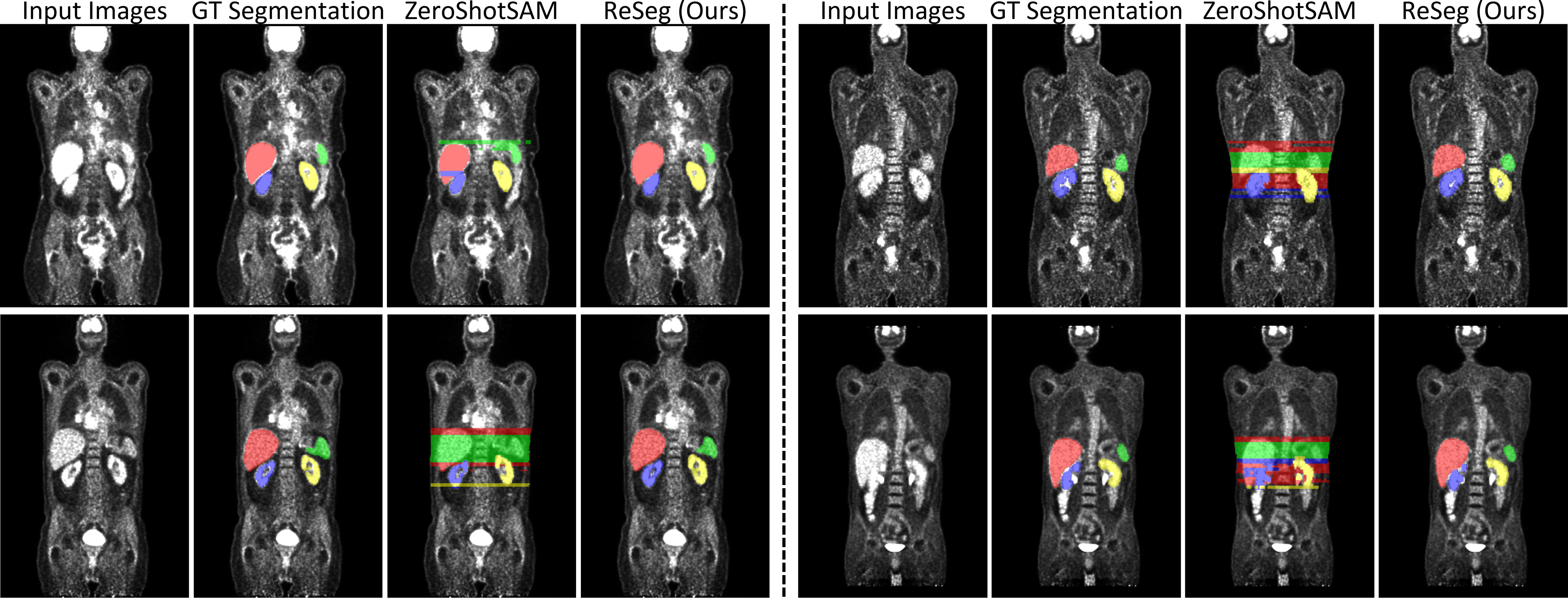}
\caption{Visual comparison results for abdominal segmentation on PET images. 3D examples from the AutoPET-Organ dataset are shown in 2D slices along the frontal plane. Left to right: input images, GT segmentation labels, and segmentation results of ZeroShotSAM \citep{shaharabany2024zero} and the proposed ReSeg. Colors correspond to the organs as follows: red - the liver, green - the spleen, blue - the right kidney, yellow - the left kidney.}
\label{fig_compare_PET}
\end{figure}

\subsubsection{Results on the generalization evaluation datasets}
We further assessed the segmentation performance of ReSeg against zero-shot SOTA methods on the three generalization evaluation datasets across CT, MRI, and PET modalities. As presented in Table \ref{tab_compare_zeroshot}, ReSeg demonstrated superior generalization capability compared to other zero-shot methods across all the three modalities. Concretely, on the WORD, the MRISegmentator, and the AutoPET-Organ datasets, ReSeg achieved respectively an average DSC of 89.62\%, 90.04\%, and 72.40\%, demonstrating substantial DSC improvements of 5.37\%, 7.67\%, and 55.72\% over the second-best method (ZeroShotSAM~\citep{shaharabany2024zero}). Notably, ReSeg operates without requiring any finetuning or prompts, whereas ZeroShotSAM necessitates test-time finetuning and foreground prompt points. 

It is worth pointing out that when evaluated on PET images, all other zero-shot methods failed to produce meaningful results, while ReSeg generated reasonable segmentation with an average DSC of 82.81\% for the liver, 69.06\% for the spleen, 68.49\% for the right kidney, and 69.24\% for the left kidney. Fig. \ref{fig_compare_PET} further visualizes the segmentation results on the AutoPET-Organ dataset. It can be observed that ReSeg accurately delineates boundaries of the target organs even in the presence of significant noise in the PET images.

\subsection{Ablation study}
We conducted comprehensive ablation studies on the four core evaluation datasets (BTCV, AMOS-CT, L2RAb-MR, and AMOS-MR): 1) We first investigated the impact of the dynamic lower bound strategy of the proposed GA model on segmentation performance; 2) We then conducted an ablation study to compare the MCC loss for pretraining GSRNet against its variants; 3) Finally, we conducted sub-component ablations on the semantics-guided segmentation architecture of OESNet.

\begin{table}[tbp]%% placement specifier
\centering%% For centre alignment of tabular.
\caption{Quantitative results of the ablation study investigating the impact of the dynamic lower bound strategy of the proposed GA model on segmentation performance. The best results are highlighted in bold. The values in parentheses indicate the standard deviation of the DSC scores.}
\adjustbox{width=1.0\textwidth}{
\begin{tabular}{cccccc}
\toprule
\multirow{2}{*}{GA Model} & \multicolumn{2}{c}{DSC (\%) Results on CT Datasets} & \multicolumn{2}{c}{DSC (\%) Results on MR Datasets} & \multirow{2}{*}{Average DSC (\%)} \\
\cmidrule(lr){2-3} 
\cmidrule(lr){4-5} 
  & BTCV & AMOS-CT & L2RAb-MR & AMOS-MR &  \\ 
\midrule
Fixed Lower Bound $a_{\mu}$ &  87.19 (3.15) & 85.22 (5.68) & 85.10 (6.08) & 85.03 (8.82) &  85.63 (5.93)  \\
Dynamic Lower Bound $a_{\mu}$ & \textbf{88.14 (3.91)} & \textbf{88.64 (3.20)} & \textbf{86.64 (3.73)} & \textbf{90.38 (1.87)} & \textbf{88.45 (3.18)} \\
\bottomrule
\end{tabular}
}
\label{tab_abla_image_syn}
\end{table}

\subsubsection{Results on the impact of the dynamic lower bound strategy of the GA model}
When designing the data engines, we introduce the GA model, which employs a dynamic lower bound strategy to enhance variation in image contrast, thereby improving model generalization capability. We examined the dynamic lower bound strategy in the GA model by replacing it with a fixed lower bound as done in Anatomix \citep{dey2025learning}, where $a_{\mu}$ is set to $a_{\mu}^L$ instead of sampling from $U(a_{\mu}^L,a_{\mu}^H)$. Quantitative analysis in Table \ref{tab_abla_image_syn} reveals consistent performance drops when using the fixed lower bound alternative: an average DSC drop of $1.54\%$, $5.35\%$, $0.95\%$, and $3.42\%$ on the L2RAb-MR, the AMOS-MR, the BTCV, and the AMOS-CT datasets, respectively. These results confirm that our intensity randomization approach effectively enhances model robustness to contrast variations.

\begin{table}[tbp]%% placement specifier
\centering%% For centre alignment of tabular.
\caption{Quantitative results of the ablation study comparing the MCC loss for pretraining GSRNet against its variants. The best results are highlighted in bold. The values in parentheses indicate the standard deviation of the DSC scores.}
\adjustbox{width=1.0\textwidth}{
\begin{tabular}{cccccccc}
\toprule
\multirow{2}{*}{\makecell[c]{AS \\ Decoder}} & \multicolumn{2}{c}{CS Decoder} & \multicolumn{2}{c}{DSC (\%) Results on CT Datasets} & \multicolumn{2}{c}{DSC (\%) Results on MR Datasets} & \multirow{2}{*}{Average DSC (\%)} \\
\cmidrule(lr){2-3} 
\cmidrule(lr){4-5} 
\cmidrule(lr){6-7} 
 & Training at Image Level & Training at Patch Level & BTCV & AMOS-CT & L2RAb-MR & AMOS-MR &   \\ 
\midrule
\ding{51} &   & &  87.38 (3.28) & 87.44 (4.56) & 83.73 (7.35) & 87.42 (8.03) & 86.49 (5.81)   \\
\ding{51} & \ding{51} &  &  87.70 (2.79) & 87.51 (7.65) & 85.07 (3.59) & 88.66 (6.17) & 87.24 (5.05) \\
\ding{51} & \ding{51} & \ding{51} & \textbf{88.14 (3.91)} & \textbf{88.64 (3.20)} & \textbf{86.64 (3.73)} & \textbf{90.38 (1.87)} & \textbf{88.45 (3.18)} \\
\bottomrule
\end{tabular}
}
\label{tab_abla_voxel_contras}
\end{table}

\subsubsection{Results of comparing the MCC loss against its variants}
In the pretraining stage, GSRNet is optimized by a novel MCC loss that encourages domain-invariant feature alignment amidst contextual variations through a CS decoder. To verify the effectiveness of the context-aware contrastive strategy, we compared the proposed MCC loss against two ablation variants that did not consider contextual consistency during pretraining:
\begin{itemize}
\item \textbf{AS decoder-only variant}: The GSRNet only had the AS decoder trained by the DPC loss.
\item \textbf{Dual-decoder variant}: The GSRNet had the AS decoder trained by the DPC loss, and another decoder trained by a voxel-wise contrastive loss using the original image pairs, which is the loss used in Anatomix \citep{dey2025learning}.
\end{itemize}

Results of this ablation study are summarized in Table \ref{tab_abla_voxel_contras}. From this table, one can see that our multi-patch context-aware contrastive strategy outperformed both ablation variants, achieving an average DSC of $88.45\%$ across the four abdominal datasets. This represents an average DSC improvement of $1.96\%$ ($P<0.001$) and $1.21\%$ ($P<0.001$) over the AS decoder-only variant (average DSC: $86.49\%$) and the dual-decoder variant (average DSC: $87.24\%$), respectively. These findings demonstrate that our MCC loss effectively enforces contextual consistency from contrastive learning among cropped pseudo-image patches, thereby improving the generalization capability of the learned features.

\begin{table}[tbp]%% placement specifier
\centering%% For centre alignment of tabular.
\caption{Quantitative results of sub-component ablations on the semantics-guided segmentation architecture of OESNet. The best results are highlighted in bold. The values in parentheses indicate the standard deviation of the DSC scores.}
\adjustbox{width=1.0\textwidth}{
\begin{tabular}{cccccccc}
\toprule
\multirow{2}{*}{\makecell[c]{CNN \\ Encoder}} & \multirow{2}{*}{\makecell[c]{Pretrained Encoder \\ Weight Initialization}} & \multirow{2}{*}{SFE} & \multicolumn{2}{c}{DSC (\%) Results on CT Datasets} & \multicolumn{2}{c}{DSC (\%) Results on MR Datasets} & \multirow{2}{*}{Average DSC (\%)} \\
\cmidrule(lr){4-5} 
\cmidrule(lr){6-7} 
 & &  & BTCV & AMOS-CT & L2RAb-MR & AMOS-MR &  \\ 
\midrule
\ding{51} & & & 85.54 (3.38) & 78.98 (14.54) & 83.86 (6.41) & 89.03 (2.11) &  84.35 (6.61)   \\
\ding{51} &\ding{51} &  & 84.44 (5.76) & 86.05 (5.98) & 85.68 (3.97) & 90.23 (1.83) &  86.60 (4.38)   \\
\ding{51} &\ding{51} & \ding{51} & \textbf{88.14 (3.91)} & \textbf{88.64 (3.20)} & \textbf{86.64 (3.73)} & \textbf{90.38 (1.87)} & \textbf{88.45 (3.18)} \\
\bottomrule
\end{tabular}
}
\label{tab_abla_network}
\end{table}

\subsubsection{Results of sub-component ablations on the semantics-guided segmentation architecture of OESNet}
To gain insights into the efficacy of the semantics-guided segmentation architecture of the proposed OESNet, we conducted comprehensive investigations of its sub-components, i.e., the pretrained encoder weight initialization and the SFE module. Through systematic ablation studies, we evaluated the effects of:
\begin{itemize}
\item \textbf{Pretrained encoder weight initialization:} Random initialization vs. transfer learning from the pretrained encoder weights.
\item \textbf{Semantics-guided Feature Extraction (SFE):} Standard convolutional blocks (two $3\times3\times3$ convolutional layers) vs. our proposed SFE module.
\end{itemize}

Table \ref{tab_abla_network} presents the results of this ablation study. From this table, one can observe a progressive performance degradation when cumulatively removing the key components mentioned above. Specifically, removal of the SFE modules resulted in an average DSC of $86.60\%$, indicating a $1.85\%$ performance degradation ($P<0.001$). Significant deterioration occurred when simultaneously disabling pretrained encoder weight initialization and the SFE modules, yielding only $84.35\%$ DSC, with a total performance drop of $4.10\%$ ($P<0.001$). This sequential performance decline validates that all components jointly enhance generalizable segmentation capability by integrating semantic priors learned in the pretraining stage. These results validate the efficacy of our semantics-guided segmentation architecture design.

\begin{table}[tbp]%% placement specifier
\centering%% For centre alignment of tabular.
\caption{Quantitative results on the cardiac segmentation task. The best results are highlighted in bold. The values in parentheses indicate the standard deviation of the DSC scores. MYO: myocadium. LA: left atrium. RA: right atrium. LV: left ventricle. RV: right ventricle. AA: ascending aorta. PA: pulmonary artery.}
\adjustbox{width=1.0\textwidth}{
\begin{tabular}{lcccccccc}
\toprule
Method & MYO & LA & RA & LV & RV & AA & PA & Average \\ 
\midrule
\multicolumn{9}{l}{\textbf{\textit{DSC (\%) Results on the MMWHS-CT dataset}}} \\
SynthSeg \citep{billot2023synthseg} & 75.40 (8.76) & 86.84 (4.57) & 77.18 (16.85) & 84.46 (13.74) & 83.16 (5.63) & \textbf{95.90 (1.34)} & 82.50 (10.03) & 83.63 (4.35) \\
\textbf{ReSeg} \textbf{(Ours)} & \textbf{84.55 (5.93)} & \textbf{89.02 (4.62)} & \textbf{86.21 (5.29)} & \textbf{87.62 (10.08)} & \textbf{89.39 (3.83)} & 93.34 (2.60) & \textbf{84.20 (7.64)} & \textbf{87.76 (2.51)} \\
\midrule
\multicolumn{9}{l}{\textbf{\textit{DSC (\%) Results on the MMWHS-MR dataset}}} \\
SynthSeg \citep{billot2023synthseg} & \textbf{81.99 (1.97)} & 86.02 (4.48) & 69.62 (24.33) & \textbf{91.47 (1.17)} & 75.88 (11.45) & 83.04 (7.36) & 77.82 (13.07) & 80.83 (7.09) \\
\textbf{ReSeg} \textbf{(Ours)} & 77.36 (4.76) & \textbf{87.87 (3.55)} & \textbf{83.62 (6.25)} & 90.52 (1.16) & \textbf{83.46 (2.03)} & \textbf{85.08 (6.44)} & \textbf{83.05 (5.46)} & \textbf{84.42 (2.72)} \\
\bottomrule
\end{tabular}
}
\label{tab_cardiac}
\end{table}

\begin{figure}[tbp]
\centering%% For centre alignment of image.
\includegraphics[width=1.0\textwidth]{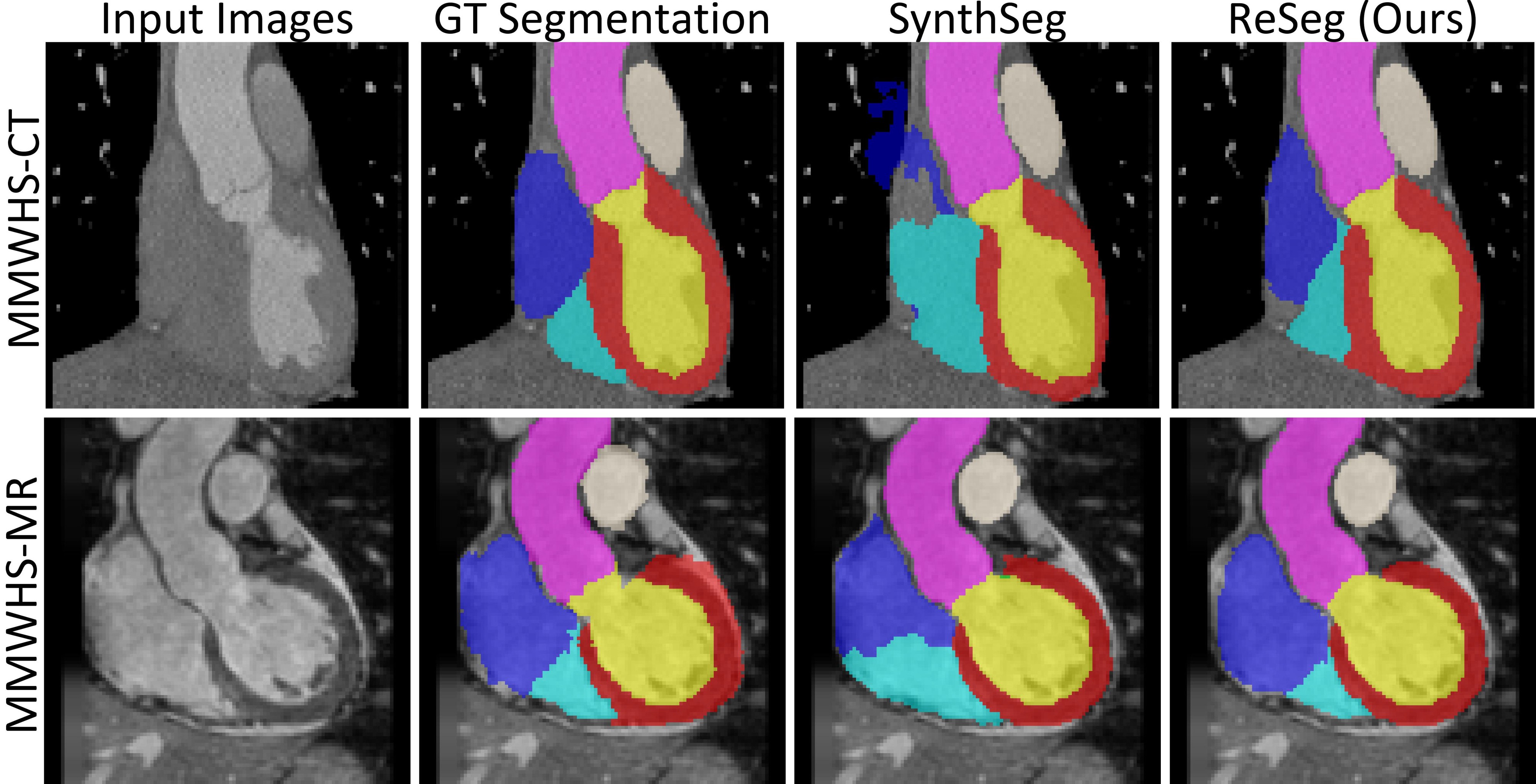}
\caption{Visual comparison results for cardiac segmentation. 3D examples from the MMWHS-CT and the MMWHS-MR datasets are shown in 2D slices along the frontal plane. Left to right: input images, GT segmentation labels, and segmentation results of SynthSeg \citep{billot2023synthseg} and the proposed ReSeg. Colors correspond to the organs as follows: red - the myocadium, green - the left atrium, blue - the right atrium, yellow - the left ventricle, cyan - the right ventricle, purple - the ascending aorta, beige - the pulmonary artery.}
\label{fig_compare_cardiac}
\end{figure}

\subsection{Extension to cardiac segmentation}
To demonstrate the generalizability of ReSeg, we applied it to the cardiac segmentation task. The study was conducted on the MMWHS dataset \citep{zhuang2019evaluation}, which comprises 20 MR scans (MMWHS-MR) and 20 CT scans (MMWHS-CT), each annotated with manual labels for seven cardiac-related structures (as detailed in Table \ref{tab_cardiac}). Following the experimental setup introduced in \cite{billot2023synthseg}, we randomly selected 13 label maps from MMWHS-MR for synthesizing training dataset. We then randomly chose two additional scans (one from the MMWHS-MR scans and the other from the MMWHS-CT scans) for validation, and all remaining scans from both MMWHS-MR and MMWHS-CT were used for testing. For preprocessing, all training label maps and test scans were resampled to an isotropic resolution of $1.0\times1.0\times1.0~\mathrm{mm}^3$ and cropped to a uniform volume size of $160\times160\times160$ voxels.

We adapted ReSeg for cardiac segmentation using the cardiac training label maps described above. The pretraining stage, which optimized GSRNet, remained fixed due to its universal representation capacity acquired through organ-agnostic semantic representation learning. In the segmentation learning stage, we maintained the overall architecture but modified the Label Processing (LP) model within the GenSeg data engine and retrained OESNet using pseudo cardiac data. Specifically, in the revised LP model, the original training label maps, which delineated only the target cardiac structures, were refined by subdividing background and foreground regions into finer tissue subcategories, following the procedure as introduced by \cite{billot2023synthseg}. These refined anatomical label maps were then used for intensity sampling to synthesize training data. For each anatomical label map, we generated 1200 offline synthetic images. Training supervision was provided by the original training label maps. We adopted a patch size of $160\times160\times160$ voxels at $1.0\times1.0\times1.0~\mathrm{mm}^3$ spacing for our method, while keeping all other hyperparameters consistent with our core experimental setup. During inference, ReSeg segmented real scans as described in Section \ref{subsec_inference}, without retraining or finetuning.

We compared ReSeg against SynthSeg \citep{billot2023synthseg}. For fair comparison, we reimplemented SynthSeg using its official codebase under our experimental setting, as our detailed data partitioning and preprocessing protocols might be different from those used in the original SynthSeg paper. Quantitative performance comparisons are summarized in Table \ref{tab_cardiac}. As shown in this table, ReSeg outperformed SynthSeg on both MMWHS-CT and MMWHS-MR datasets. Specifically, on the MMWHS-CT dataset, ReSeg achieved an average DSC of 87.76\%, representing a 4.13\% improvement over SynthSeg. On the MMWHS-MR dataset, it attained an average DSC of 84.42\%, outperforming SynthSeg by 3.59\%. Fig. \ref{fig_compare_cardiac} presents the visual comparison results on these two datasets. It can be seen that ReSeg produces more anatomically accurate cardiac segmentation compared with SynthSeg. These results indicate the generalizability of ReSeg for organ segmentation tasks beyond abdominal organs.

\section{Discussion}

Domain-randomized image synthesis offers a promising way to eliminate the need for real training images and support generalizable segmentation learning. However, it remains underexplored whether such synthetic data-driven training paradigms can be effectively applied to generalizable abdominal organ segmentation without requiring domain-specific adaptation. To mitigate the gap, we propose ReSeg, a two-stage retraining-free framework for zero-shot segmentation of abdominal scans. Built upon two delicately designed data engines, our proposed two-stage network first learns organ-agnostic, generalizable semantic features, and then transitions to acquiring abdominal-organ-specific semantic features. We conducted comprehensive experiments on seven abdominal segmentation datasets encompassing different imaging modalities (CT, MRI, and PET), varying spatial resolutions, and diverse patient demographics to evaluate the generalization capability of ReSeg to various unseen domains. Across all evaluation datasets, our method consistently outperformed the competing SOTA approaches, as evidenced by the results presented in Tables \ref{tab_compare_ct}, \ref{tab_compare_mr}, and \ref{tab_compare_zeroshot}, and Figures \ref{fig_compare_ct}, \ref{fig_compare_mr}, \ref{fig_compare_box}, and \ref{fig_compare_PET}. To further investigate the effectiveness of the proposed GA model, the MCC loss, and the architectural components of the proposed OESNet, we conducted comprehensive ablation studies. The experimental results presented in Tables \ref{tab_abla_image_syn}, \ref{tab_abla_voxel_contras}, and \ref{tab_abla_network} provide a detailed assessment of how different components of the proposed method contribute to the segmentation generalization performance.

Direct application of existing domain-randomization-based segmentation training paradigms like SynthSeg \citep{billot2023synthseg} to abdominal organ segmentation proved ineffective (Tables \ref{tab_compare_ct}, \ref{tab_compare_mr}, and \ref{tab_compare_zeroshot}). Thus, we develop two dedicated data engines with the newly designed GA model to enhance feature learning in the complex scenarios of volumetric abdominal images. In particular, the GA model uses a dynamic lower bound for the mean parameters in the GMMs to better capture the large contrast variations encountered in abdominal imaging. This strategy significantly enhances the generalization capability of the segmentation network, as evidenced by the improved performance shown in Table \ref{tab_abla_image_syn}. Our delicately designed data engines enable generalizable segmentation learning for abdominal anatomy, which exhibit high variations in appearance and background among patients and thus pose significant zero-shot segmentation challenges.

Built upon the data engines, the key to the superior generalization performance of ReSeg lies in its two-stage learning framework, which first acquires generalizable semantic representations for diverse anatomies and then learns organ-specific features guided by the learned semantic priors. The generalizable semantic representation learning stage utilizes the GenRep data engine, which produces paired synthetic images with diverse semantic layouts from randomized label maps. Then, GSRNet is trained via a novel MCC loss that encourages domain-invariant feature alignment amidst contextual variations. This approach effectively learns domain-invariant context-aware semantic features by leveraging multiple overlapping pseudo-image patches. By comparison, the voxel-wise contrastive loss as proposed by \cite{dey2025learning}, which relies solely on pseudo-image pairs, learns appearance-agnostic features but overlooks contextual consistency, leading to diminished performance when segmenting images with contextual changes across varying FOVs, as confirmed by the results presented in Table \ref{tab_abla_voxel_contras}. The subsequent segmentation training employs the GenSeg data engine, which generates synthetic images with anatomical spatial layouts by preserving organ position arrangements, enabling effective learning of abdominal-organ-specific features. To gain robustness against abdominal imaging variations, OESNet incorporates rich semantic priors derived from the well-trained GSRNet to guide feature extraction within the SS decoder. The efficacy of such a design is demonstrated by the results presented in Table \ref{tab_abla_network}.

The proposed ReSeg framework extends beyond abdominal organ segmentation and can be adapted to other medical image segmentation tasks. The pretraining stage learns organ-agnostic semantic features, which optimizes GSRNet on semantically and visually diverse synthetic images, using loss functions that focus on appearance invariance and contextual consistency. This enables the network to acquire semantic features that are robust against variations in anatomical structure, imaging contrast, resolution, and spatial context, thereby establishing a universal representational foundation for segmentation learning. In the subsequent segmentation learning stage, the overall semantics-guided segmentation architecture can be applied to new anatomical targets without being restricted to abdominal organs. In order to adapt ReSeg for new segmentation tasks, it is sufficient to modify the LP model within the dedicated data engine for segmentation learning and then retrain OESNet. Experimental results presented in Table \ref{tab_cardiac} confirm the effectiveness of extending ReSeg to the cardiac segmentation task, demonstrating the generalizability of ReSeg for organ segmentation beyond abdominal organs.

Compared to existing medical image segmentation paradigms \citep{he2025vista3d,billot2023synthseg,dey2025learning}, the proposed method offers several advantages. First, while most existing biomedical foundation segmentation models \citep{liu2023clip,ma2024segment,he2025vista3d} rely heavily on large-scale medical datasets covering multiple imaging domains to achieve generalization, our proposed method benefits from the synthetic data-driven training strategy based on domain randomization, eliminating completely the requirement of acquiring any real training image. Second, unlike existing domain-randomization-based segmentation training paradigms \citep{billot2023synthseg,chollet2024label,dey2024anystar} designed for structures with relatively fixed anatomical topology and stable tissue intensity distribution, which yield limited segmentation performance when applied to abdominal organs, the proposed ReSeg achieved superior abdominal organ segmentation performance, as evidenced by the results presented in Tables \ref{tab_compare_ct}, \ref{tab_compare_mr}, and \ref{tab_compare_zeroshot}. Third, unlike the previously introduced Anatomix \citep{dey2025learning}, which is pretrained via contrastive learning on paired pseudo-images, neglecting contextual consistency and requiring downstream finetuning on real medical images, the proposed ReSeg combines organ-agnostic pretraining with organ-specific segmentation training, enabling retraining-free abdominal organ segmentation, which is a clear advantage for clinical deployment. Finally, when it comes to generalizable abdominal organ segmentation, existing zero-shot segmentation methods suffer from limited generalization capabilities, particularly under significant domain shifts, as evidenced by the experimental results presented in Tables \ref{tab_compare_ct}, \ref{tab_compare_mr}, and \ref{tab_compare_zeroshot}. In contrast, our ReSeg demonstrates reliable segmentation across multiple imaging modalities, even when it is trained only on synthetic data. Notably, despite complete exclusion of real clinical images during training, ReSeg achieves superior segmentation performance against few-shot SOTA methods, including few-shot meta-learning approaches \citep{ouyang2022self,tang2025few,cheng2025dual} and finetuned representation learning models \citep{wu2024voco,dey2025learning}, all of which require access to annotated real data.

\section{Conclusion}
This paper introduced ReSeg, a novel two-stage retraining-free framework designed to achieve generalizable segmentation of abdominal organs. Specifically, at the first stage, an organ-agnostic semantic representation learning strategy was proposed, which exploited semantically and visually diverse synthetic images generated by a delicately designed data engine to extract semantic features robust against variations in imaging appearance and spatial context. Such strategy endowed the pretrained network with universal representation capabilities, laying a solid foundation for subsequent segmentation learning. After that, at the second stage, a semantics-guided segmentation architecture was introduced, which was optimized using diversified visual images with anatomical spatial layouts synthesized by another delicately designed data engine. Within the segmentation network, rich semantic priors derived from the pretrained network were hierarchically integrated into the decoder, enabling effective learning of abdominal-organ-specific semantic features. Comprehensive validation on seven abdominal datasets across CT, MRI, and PET modalities demonstrated the superior generalization capability of ReSeg compared to SOTA methods.

\section*{CRediT authorship contribution statement}
\textbf{Derong Yu:} Writing - original draft, Visualization, Validation, Software, Methodology, Investigation, Data curation, Conceptualization, Formal analysis; \textbf{Guoyan Zheng:} Writing - review \& editing, Supervision, Resources, Methodology, Funding acquisition, Conceptualization.

\section*{Declaration of competing interest}
The authors declare that they have no known competing financial interests or personal relationships that could have appeared to influence the work reported in this paper.

\section*{Acknowledgements}
The work was partially supported by the National Key R\&D Program of China via project 2023YFB4706302, and by the National Natural Science Foundation of China via project 62471293. 

\bibliographystyle{elsarticle-harv} 
\bibliography{references}



\end{document}